\documentclass[a4paper,conference]{IEEEtran}
\ifCLASSINFOpdf
\else
\fi
\usepackage{amsmath}
\usepackage{algorithmic}
\usepackage{fancyhdr}
\usepackage{amsmath}
\usepackage{bm}
\usepackage{amssymb}
\usepackage{graphicx,color}
\usepackage{subfigure}
\usepackage{algorithmic,algorithm}
\usepackage{cite}
\usepackage{balance}
\usepackage{txfonts}
\usepackage{theorem}
\usepackage{here}
\usepackage{bm}
\usepackage{hhline}
\usepackage{comment}
\usepackage{multirow}
\usepackage{threeparttable}
\usepackage{breqn}
\usepackage{dcolumn}
\usepackage{bm}
\usepackage{lipsum}

\newcommand{\wfig}[1]{Figure \ref{fig:#1}}
\newcommand{\wtab}[1]{Table \ref{tab:#1}}

\begin{document}
%
\title{YuruGAN: Yuru-Chara Mascot Generator Using Generative Adversarial Networks With Clustering Small Dataset}

\author{\IEEEauthorblockN{Yuki Hagiwara}
\IEEEauthorblockA{Department of Electrical and Electronic Engineering\\
Tokyo University of Agriculture and Technology\\ 
2--24--16, Nakacho, Koganei-shi, Tokyo, 184--8588, Japan\\
Email: hagiwara19@sip.tuat.ac.jp}
\and
\IEEEauthorblockN{Toshihisa Tanaka}
\IEEEauthorblockA{Department of Electrical and Electronic Engineering\\
Tokyo University of Agriculture and Technology\\ 
2--24--16, Nakacho, Koganei-shi, Tokyo, 184--8588, Japan\\
Email: tanakat@cc.tuat.ac.jp}}


%


\maketitle

\begin{abstract}
A \textit{yuru-chara} is a mascot character created by local governments and companies for publicizing information on areas and products. Because it takes various costs to create a \textit{yuru-chara}, the utilization of machine learning techniques such as generative adversarial networks (GANs) can be expected. In recent years, it has been reported that the use of class conditions in a dataset for GANs training stabilizes learning and improves the quality of the generated images. However, it is difficult to apply class conditional GANs when the amount of original data is small and when a clear class is not given, such as a \textit{yuru-chara} image.
In this paper, we propose a class conditional GAN based on clustering and data augmentation. Specifically, first, we performed clustering based on K-means++ on the \textit{yuru-chara} image dataset and converted it into a class conditional dataset. Next, data augmentation was performed on the class conditional dataset so that the amount of data was increased five times. In addition, we built a model that incorporates ResBlock and self-attention into a network based on class conditional GAN and trained the class conditional yuru-chara dataset. As a result of evaluating the generated images, the effect on the generated images by the difference of the clustering method was confirmed.
\end{abstract}


%
\IEEEpeerreviewmaketitle

\section{Introduction}
\label{ch:introduction}
A \textit{yuru-chara} is a ``mascot character'' originating from Japan; it is typically created by local governments and companies for the purpose of promoting events, regional revitalization, products, and so forth. The number of \textit{yuru-chara}s is increasing year by year, and an annual ``the Yuru-chara Grand Prix'' ranks entries by the popularity of these \textit{yuru-charas}. The economic effects produced by \textit{yuru-chara} are significant. For example, the Kumamoto prefecture's \textit{yuru-chara} ``Kumamon,'' which won “the Yuru-chara Grand Prix 2011", has generated an economic effect of about 124.4 billion yen thanks to its popularity throughout the country\cite{kumamon}. This is one of the incentives for local governments and companies to create characters that can be ranked high in ``the Yuru-chara Grand Prix.'' Regarding the relationship between \textit{yuru-chara} and popularity, Nakasato et al. analyzed the characters that participated in ``the Yuru-chara Grand Prix'' and the correspoding rankings using machine learning and found that the design and colors of \textit{yuru-chara} greatly contributed to their popularity\cite{nakasato2017analysis}.

Because of copyrights and creator fees, it costs a lot to create a new \textit{yuru-chara}. Spending money and time to design a \textit{yuru-chara} does not guarantee its popularity.
Therefore, by automating the design of \textit{yuru-charas}, a cost reduction can be expected.

 A promising way to realize automatic image generation by machine learning is the generative adversarial networks (GANs) method proposed by Goodfellow et al\cite {Goodfellow2014}. GANs can generate images that are almost indistinguishable from training data statistically and have succeeded in generating realistic images compared with other image generative models. Furthermore, various improved versions have been proposed to improve the learning of GANs. The conditional GAN proposed by Mirza et al. reports that by giving class information to a model, a phenomenon called mode collapse\cite{Goodfellow2014, salimans2016improved} can be suppressed, and the quality of generated images can be improved\cite{mirza2014conditional}. Depending on the tuning of the model, mode collapse is a phenomenon in which a model generates only the same image. Miyato et al. also proposed a model called cGAN with projection discriminator, reporting that by devising how to give class information to the model, learning was stable and mode collapse in class units could be suppressed\cite{ miyato2018cgans}. In the self-attention GAN (SA-GAN) proposed by Zhang et al., a self-attention layer was introduced into the GAN model proposed by Miyato et al.~\cite{zhang2018self}, and this has successfully created high-quality images. 

 However, these GAN-related studies use large datasets such as ImageNet\cite{deng2009imagenet} and CIFAR-10\cite{cifar10}, as shown in \wtab{c_sample}, which are designed for machine learning research. When it comes to the datasets used in these studies, all images are preliminarily labeled with class information\cite{deng2009imagenet}. In the current paper, \textit{yuru-chara} images are used as training data; however, a total of 10,485 \textit{yuru-chara} participated in ``the Yuru-chara Grand Prix'' from 2011 to 2019 (3,908 if entries from multiple years are excluded), so the amount of data is very small compared with the research dataset, and no class label is given to the images of \textit{yuru-chara}. Therefore, there is no guarantee that the learning will be stable and that high-quality images can be generated when using loose character images as training data, even for GANs with class conditions to which various learning stabilization methods are applied.
 
\begin{table}[h]
\caption{Overview of the typical image dataset and \textit{yuru-chara} image dataset}
\centering
\begin{tabular}{|c|c|c|c|} \hline
{} & {ImageNet} & {CIFAR-10} & {\begin{tabular}{c}The Yuru-chara Grand Prix \\ (2010--2019)\end{tabular}} \\ \hline
{\begin{tabular}{c}Number \\ of \\ datas \end{tabular}} & {14,197,122}&{60,000}&{Total 10,485}\\ \hline
{\begin{tabular}{c}Number \\ of \\ classes \end{tabular}} & {21,841}&{10}&{---------------}\\ \hline
\end{tabular}
\label{tab:c_sample}
\end{table}

Therefore, in the current paper, we present how to apply class conditional GANs to the domain of a small dataset with no label. The proposed approach is based on clustering of datasets to train class conditional GANs stably and efficiently. In the proposed method, features are extracted from images using ResNet\cite{he2016deep}, and the \textit{yuru-chara} image dataset is converted into labeled data sets by performing clustering for features (either low-level or high-level) using the X-means\cite{pelleg2000x} and K-means++\cite{wagstaff2001constrained}. However, we apply data augmentation\cite{generator} to this labeled dataset and increase the number of samples by five times. For image generation, a model based on class conditional GANs is constructed. The proposed model considers a learning stabilization method and a method for improving the quality of the generated image as discussed in GAN-related works. The simulation shows that when class information obtained by various clustering methods together with a \textit{yuru-chara} image is given to an image generative model, images with a high quality and variety are stably generated without mode collapse.

\section{Method}
\subsection{Dataset}
 \subsubsection{Data Collection}
 We collected images from ``the Yuru-chara  Grand Prix.'' If a \textit{yuru-chara} was entered in more than one year, only one was added for the dataset. These \textit{yuru-chara} images have a resolution of $790 \times 650$ px.
 
 \subsubsection{Data Arrangement}
 Some of the images posted on ``the Yuru-chara Grand Prix site'' include images with backgrounds and images with tools, as shown in \wtab{dataset}. For those images, the background and tools were manually removed. Moreover some images include multiple characters. For those images, the characters were separated into single characters. For this process, if the character after separation was too small or if a part of the body was largely missing, the character was removed from the dataset. Finally, the total number of samples in the \textit{yuru-chara} image dataset after removal is 4,018. 
 
 \begin{table}[t]
\caption{Details of the \textit{\textit{yuru-chara}} dataset and images before and after processing }
\centering
\begin{tabular}{|c|c|c|c|} \hline
{} & {With background} & {With tools} & {Multiple characters} \\ \hline
{Numbers}&{247}&{380}&{36}\\  \hline
{Before}&
\begin{minipage}{18mm}
      \centering
      \includegraphics[keepaspectratio, scale=0.07]{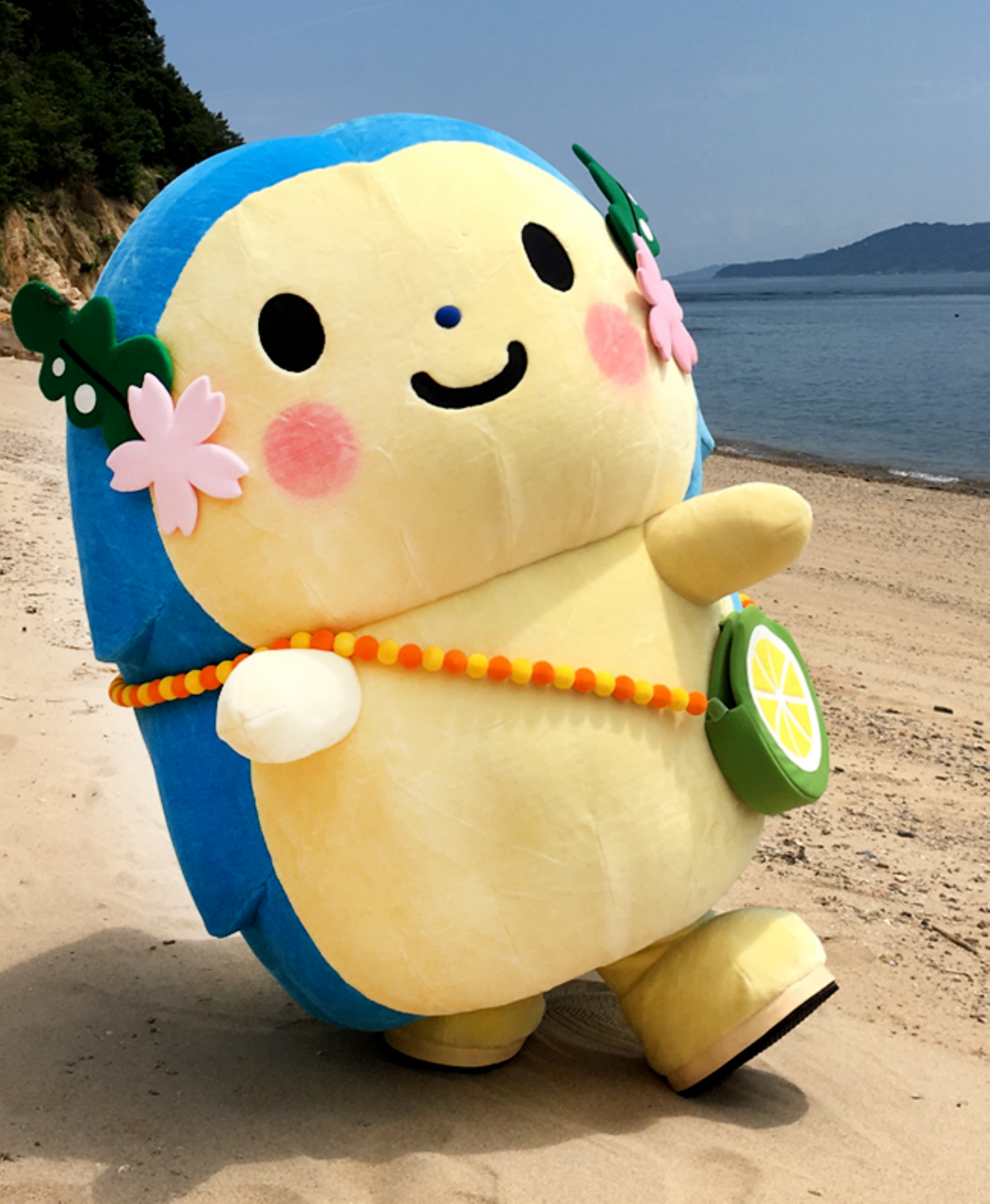}
    \end{minipage}&
\begin{minipage}{18mm}
      \centering
      \includegraphics[keepaspectratio, scale=0.07]{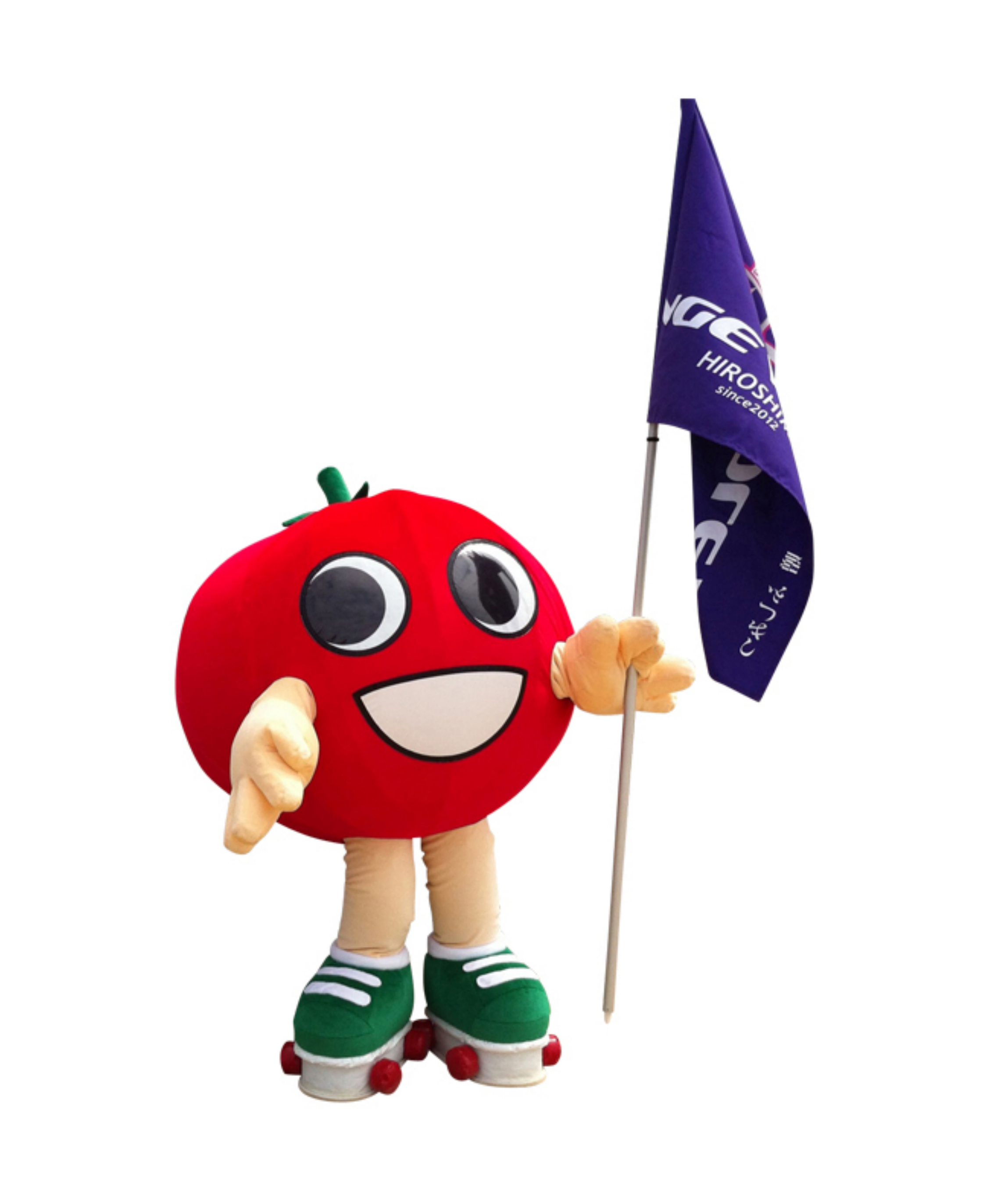}
    \end{minipage}&
\begin{minipage}{18mm}
      \centering
      \includegraphics[keepaspectratio, scale=0.07]{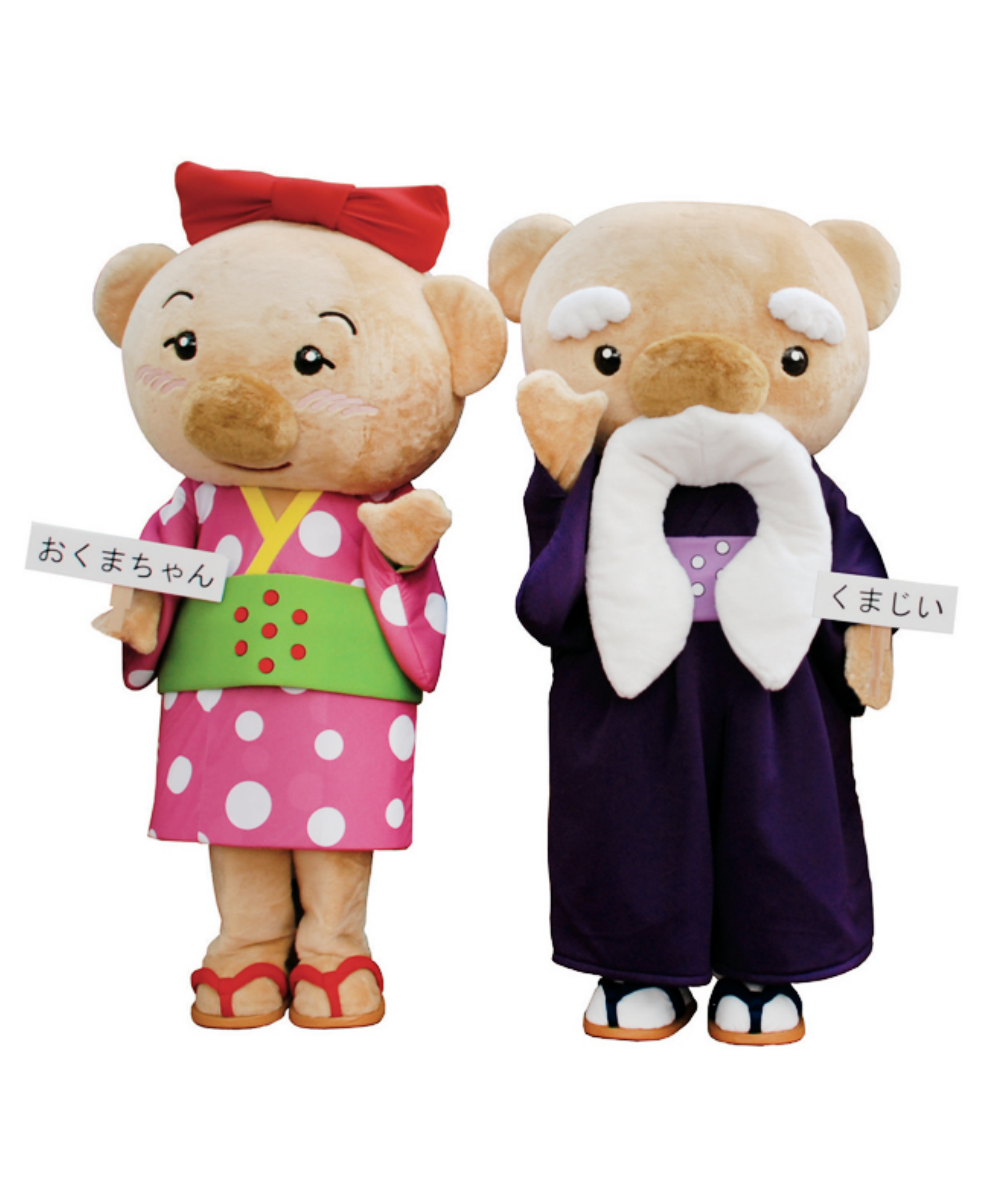}
    \end{minipage} \\ 
\hline
{After}&
\begin{minipage}{18mm}
      \centering
      \includegraphics[keepaspectratio, scale=0.07]{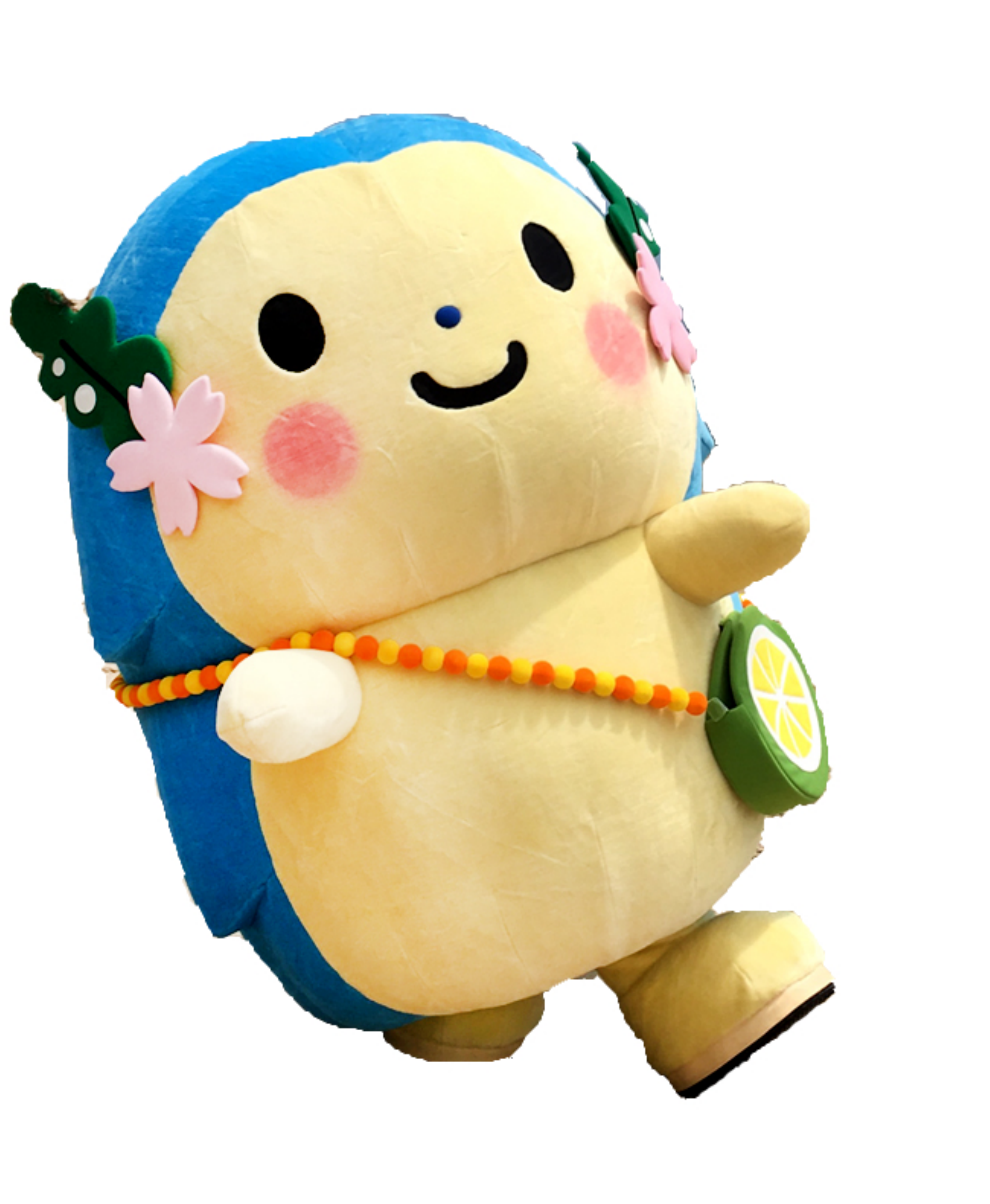}
    \end{minipage}&
\begin{minipage}{18mm}
      \centering
      \includegraphics[keepaspectratio, scale=0.07]{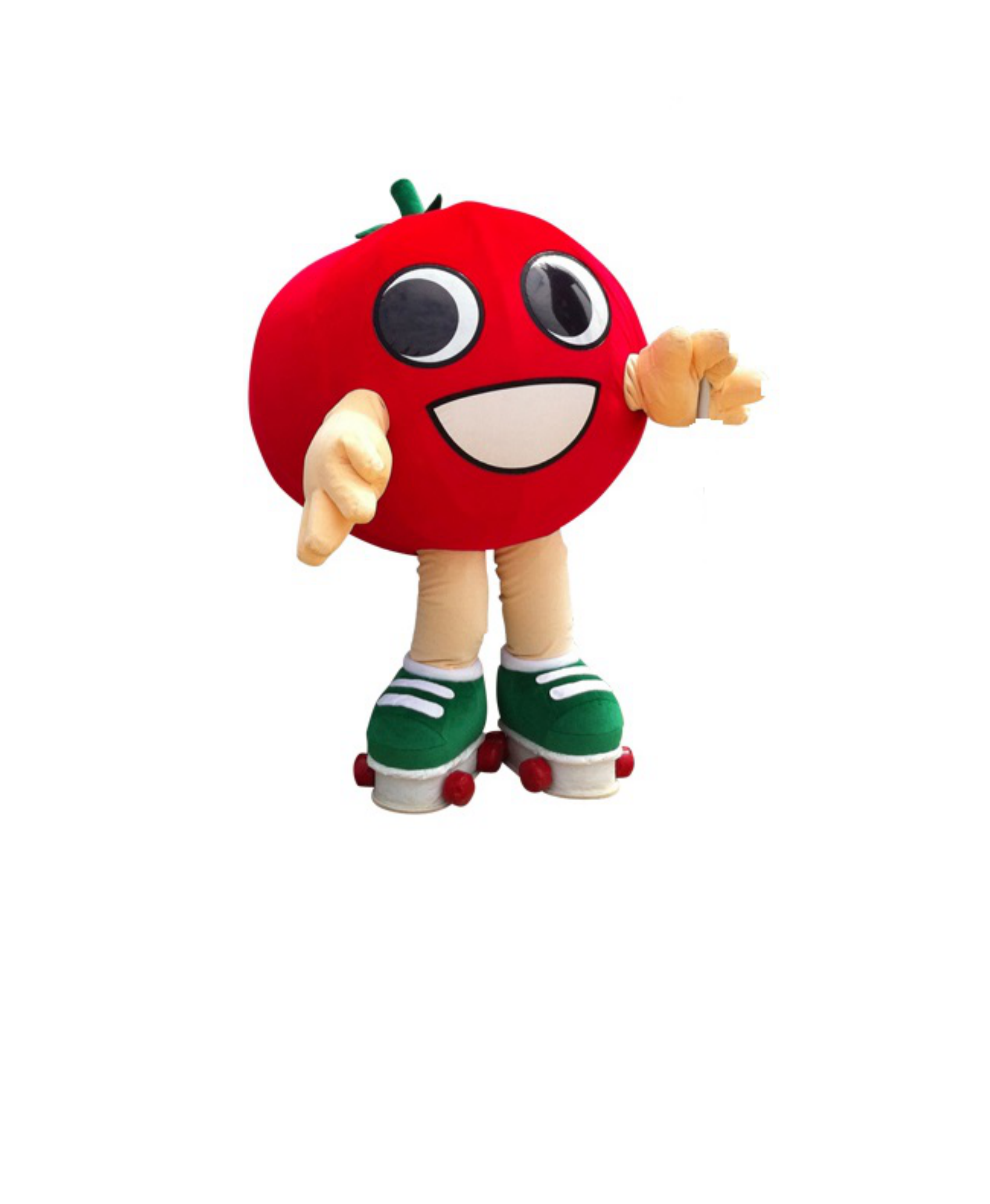}
    \end{minipage}&
\begin{minipage}{18mm}
      \centering
      \includegraphics[keepaspectratio, scale=0.07]{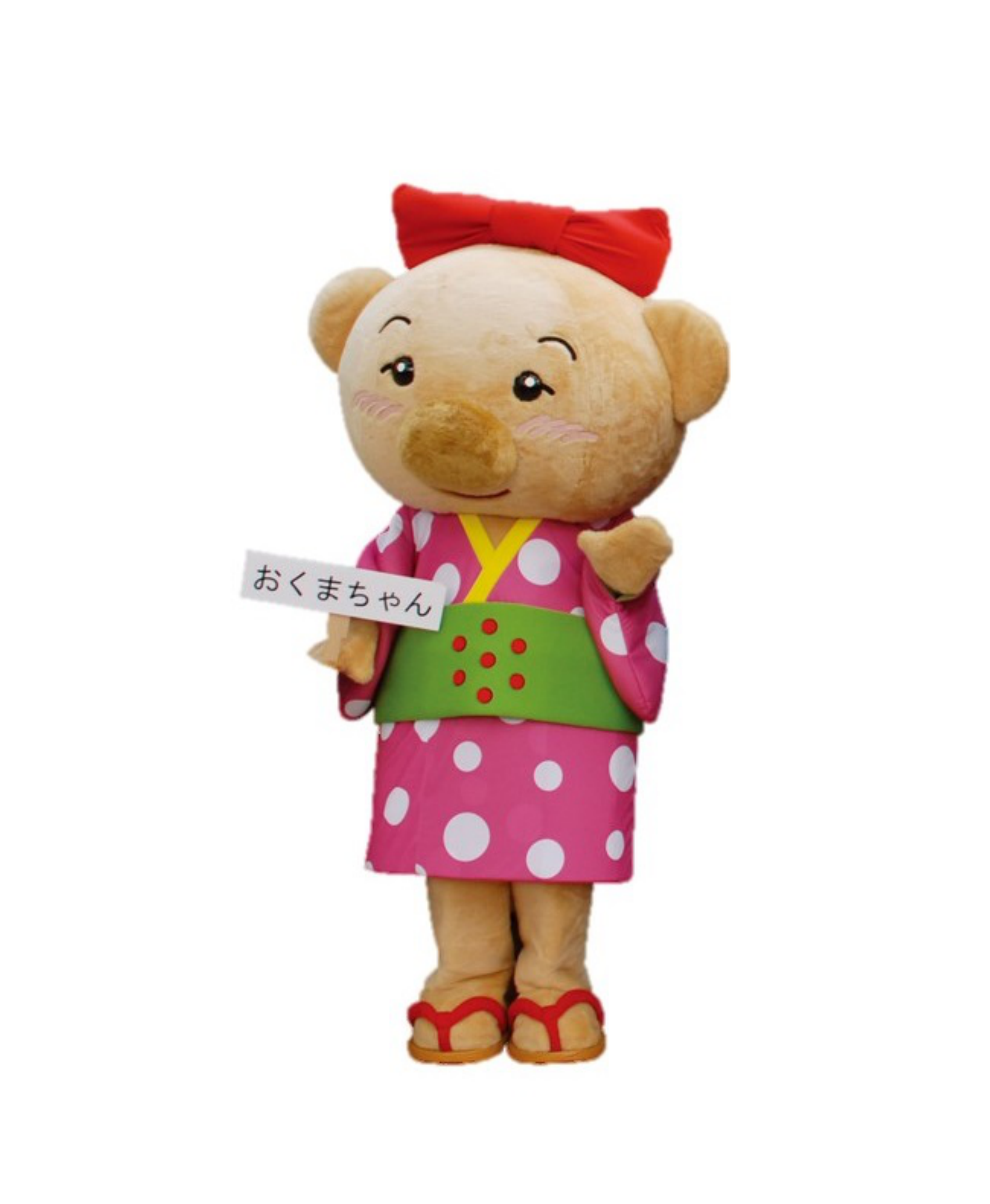}\\
      \includegraphics[keepaspectratio, scale=0.07]{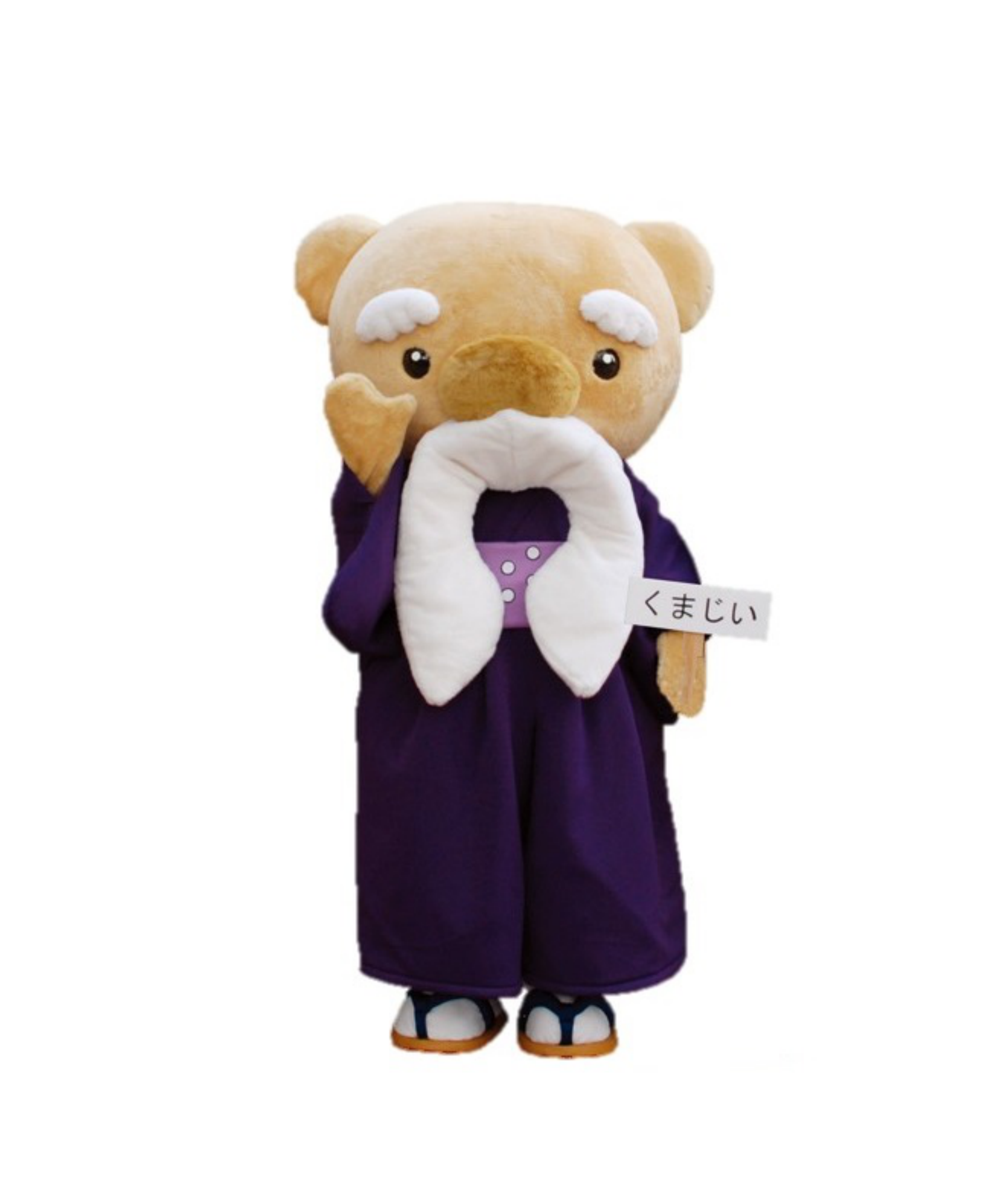}
    \end{minipage} \\ 
\hline
\end{tabular}
\label{tab:dataset}
\end{table}

\subsubsection{Resize of Data}
\begin{figure*}[h]
\begin{minipage}[t]{0.24\linewidth}
\centering
\includegraphics[keepaspectratio, scale=0.42]{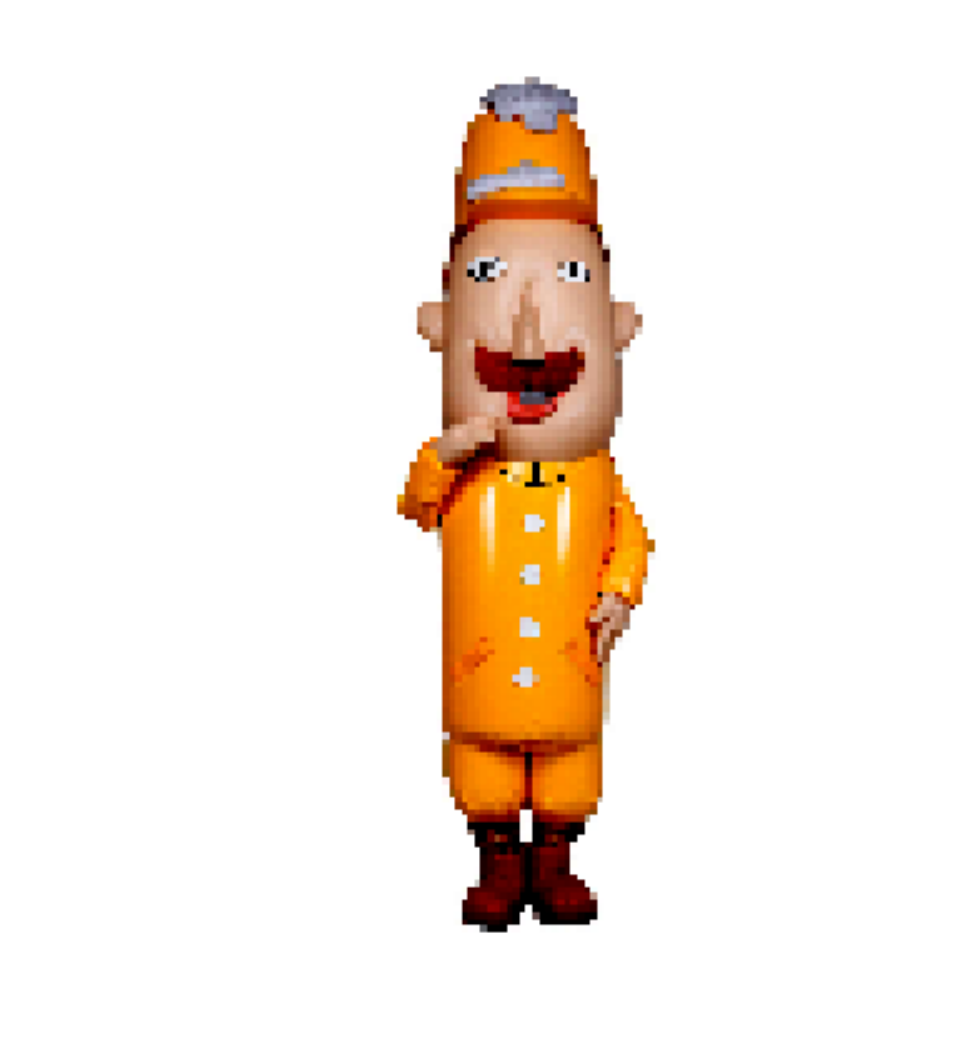}
\end{minipage}
\begin{minipage}[t]{0.24\linewidth}
\centering
\includegraphics[keepaspectratio, scale=0.42]{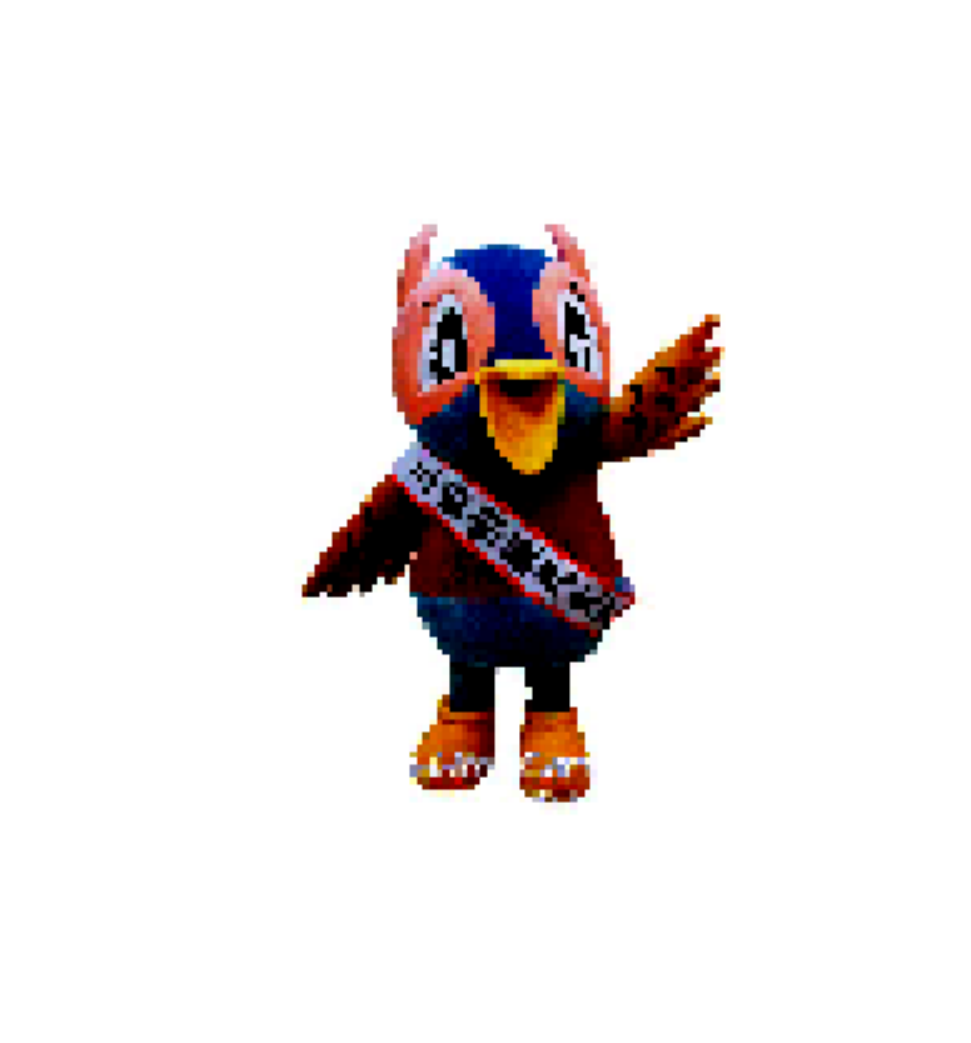}
\end{minipage}
\begin{minipage}[t]{0.24\linewidth}
\centering
\includegraphics[keepaspectratio, scale=0.42]{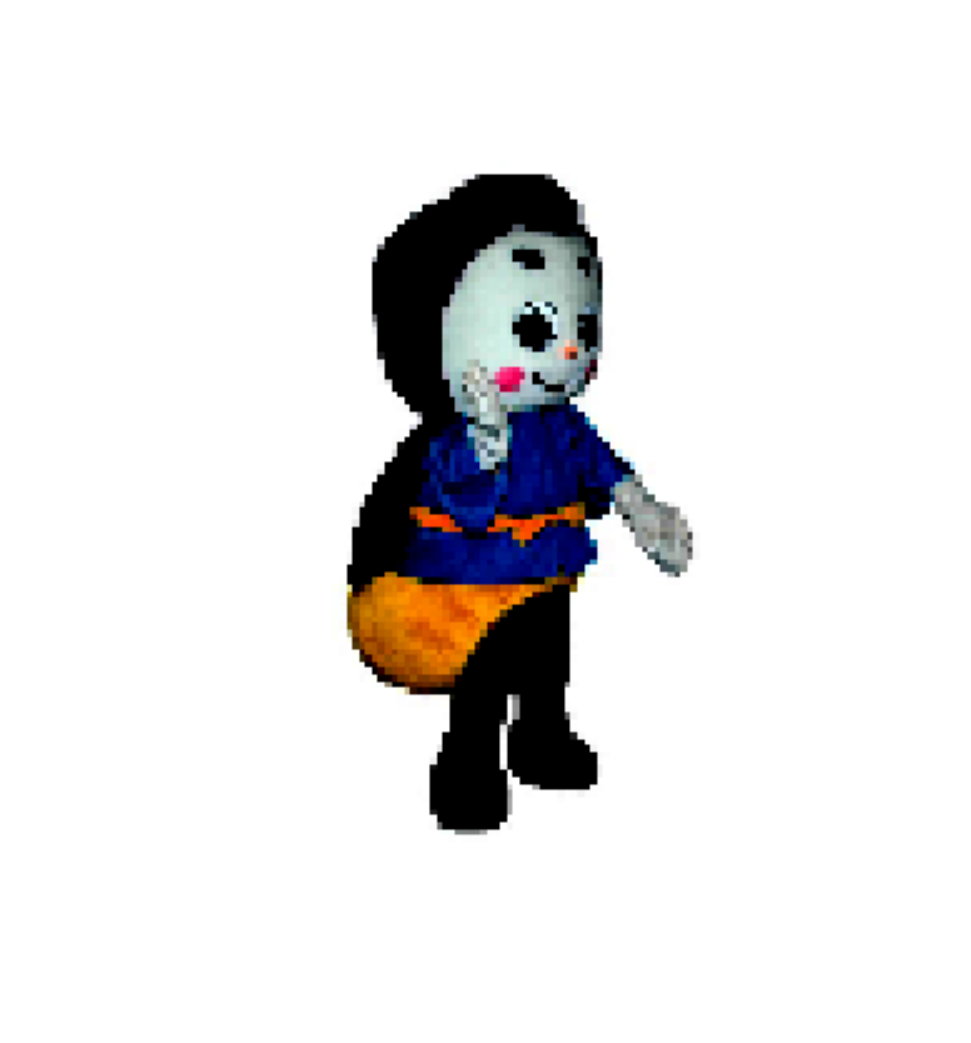}
\end{minipage}
\begin{minipage}[t]{0.24\linewidth}
\centering
\includegraphics[keepaspectratio, scale=0.42]{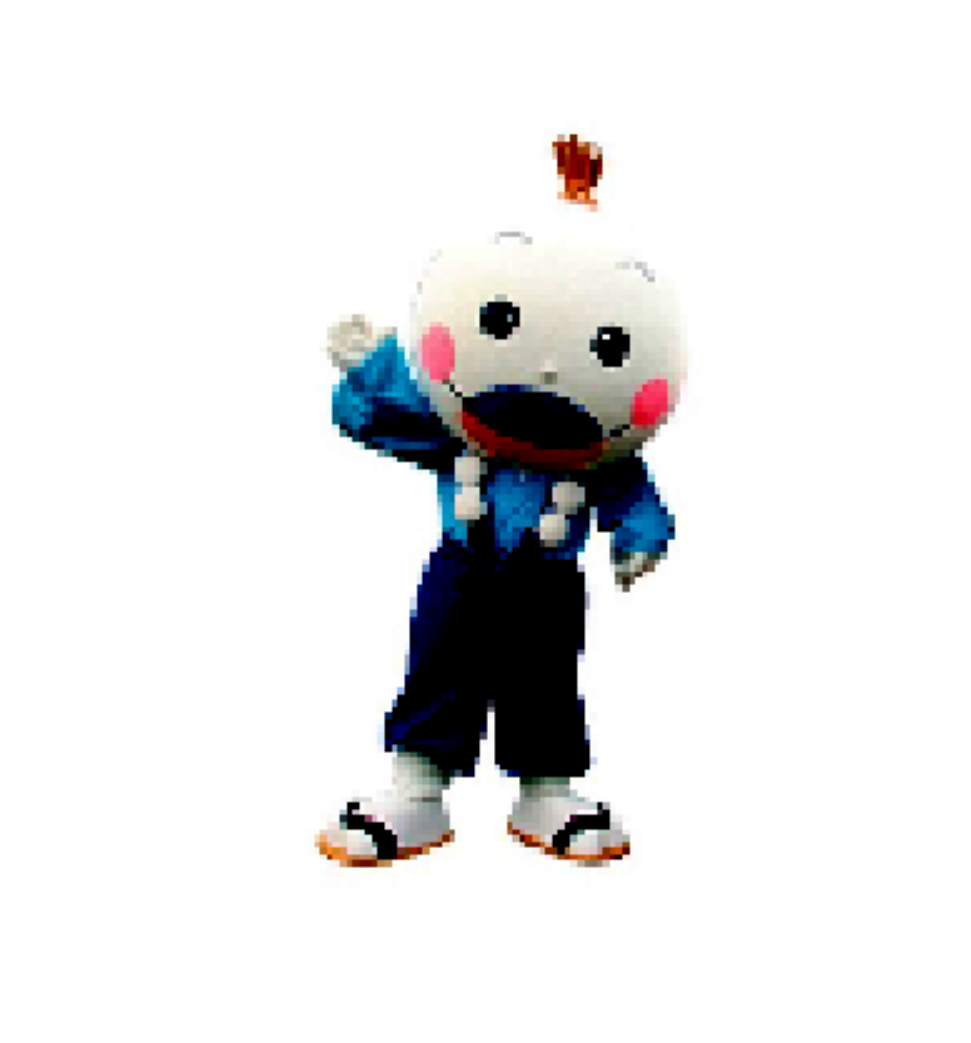}
\end{minipage}

\centering \caption{Resized samples in the \textit{basic set}.}
\label{fig:resize_sample}
\end{figure*}

As shown in \wfig{resize_sample}, we resized the arranged sample to $144 \times 128$ px. This resized dataset  consisting of 4,018 samples is named the \textit{basic set}.

 \subsubsection{Data Augmentation}
Data augmentation was applied to the \textit{basic set} of images to extend the amount of data. In this paper, we used the \texttt{ImageDataGenerator} class on Keras\cite{generator} to rotate the sample within 8 degrees left and right, slide it within 5\% left and right, and zoom the sample between 90\% and 110\% by using the sample. Data augmentation was performed so that the total number was increased by a factor of five. As a result of data augmentation, the total number of samples is 20,090. This is named the \textit{augmented set}.
 
 \subsection{Split Dataset by Clustering}
\begin{figure}[t]
　\centering
　\includegraphics[width=5.5cm]{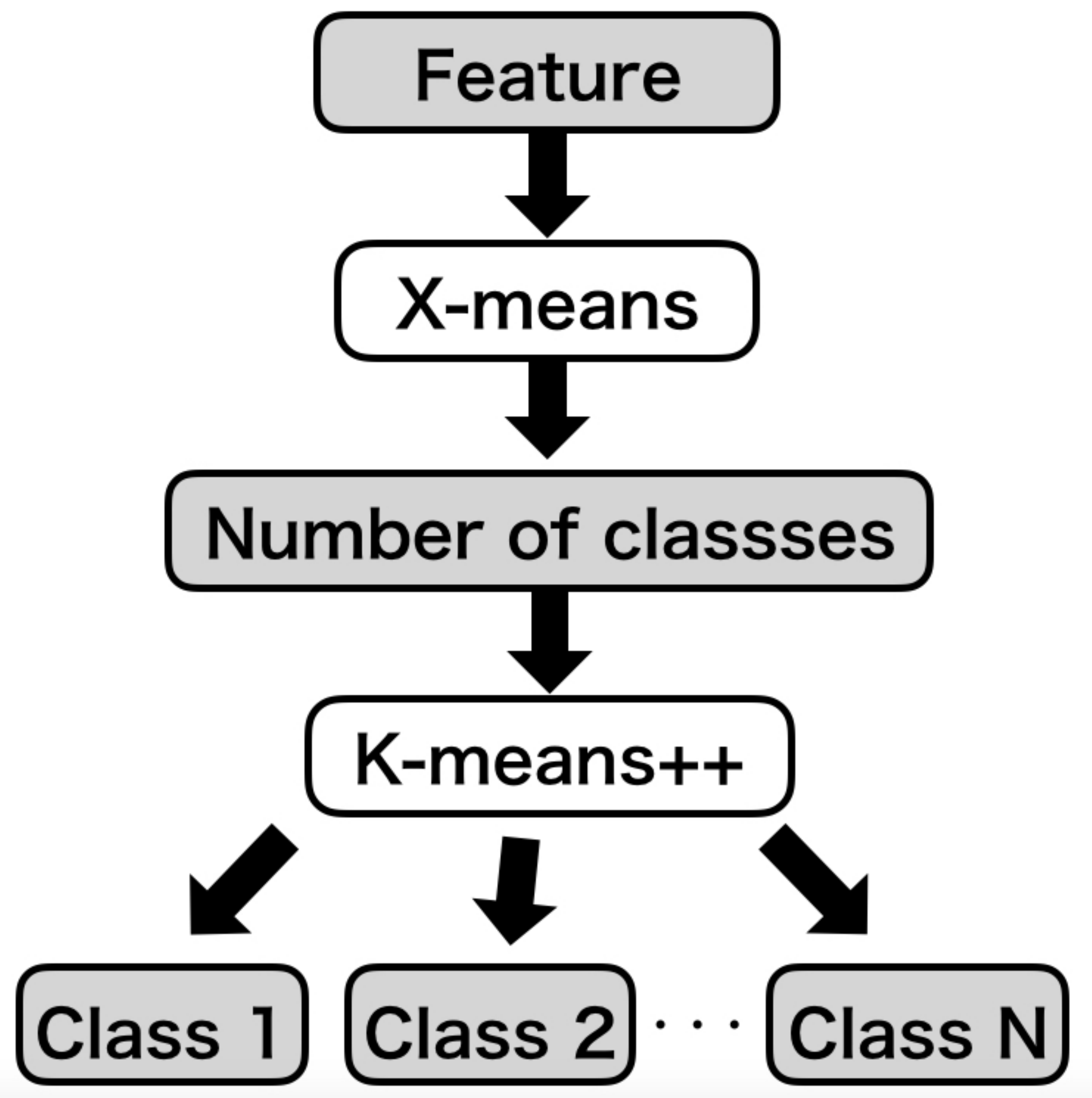}
　\caption{Clustering using X-means}
　\label{fig:clustering}
\end{figure}

To apply a class conditional GAN, we needed to convert the \textit{augmentation set} to a labeled dataset. In the current paper, the below three types of features are used in clustering. Then, X-means\cite{pelleg2000x} was applied to the features to determine the number of classes (\wfig{clustering}). The number of classes depends on the initial seed, and hence, X-means were applied ten times, and the average value was used as the number of classes.
 
 \subsubsection{RGB Feature}
 \label{sssec:no}
Color (RGB) images in the \textit{basic set} were directly applied to clustering, as shown in \wfig{rgb_clustering}. The class label obtained in this way was added to all the samples in the \textit{augmented set}, and the resulting labeled dataset is named the \texttt{RGB\_dataset}.

 \subsubsection{ResNet Feature}
 \label{sec:resnet}
\begin{figure}[t]
	\begin{tabular}{cc}
	\hspace{-15pt}
	\subfigure[RGB\_dataset]{
	\includegraphics[width = 2.9cm]{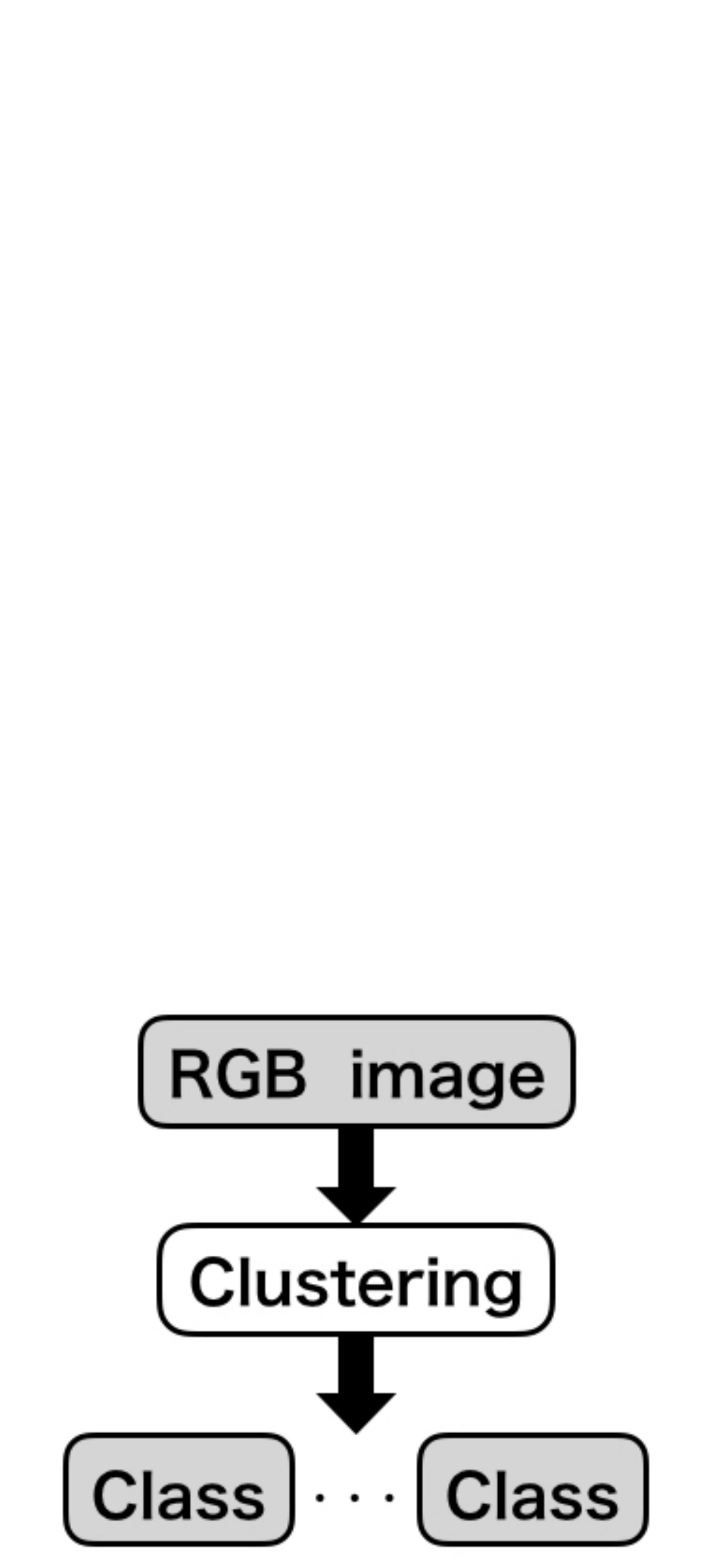}
	\label{fig:rgb_clustering}
	}
	\hspace{-13pt}
	\subfigure[RGB\_ResNet\_dataset]{
	\includegraphics[width = 2.9cm]{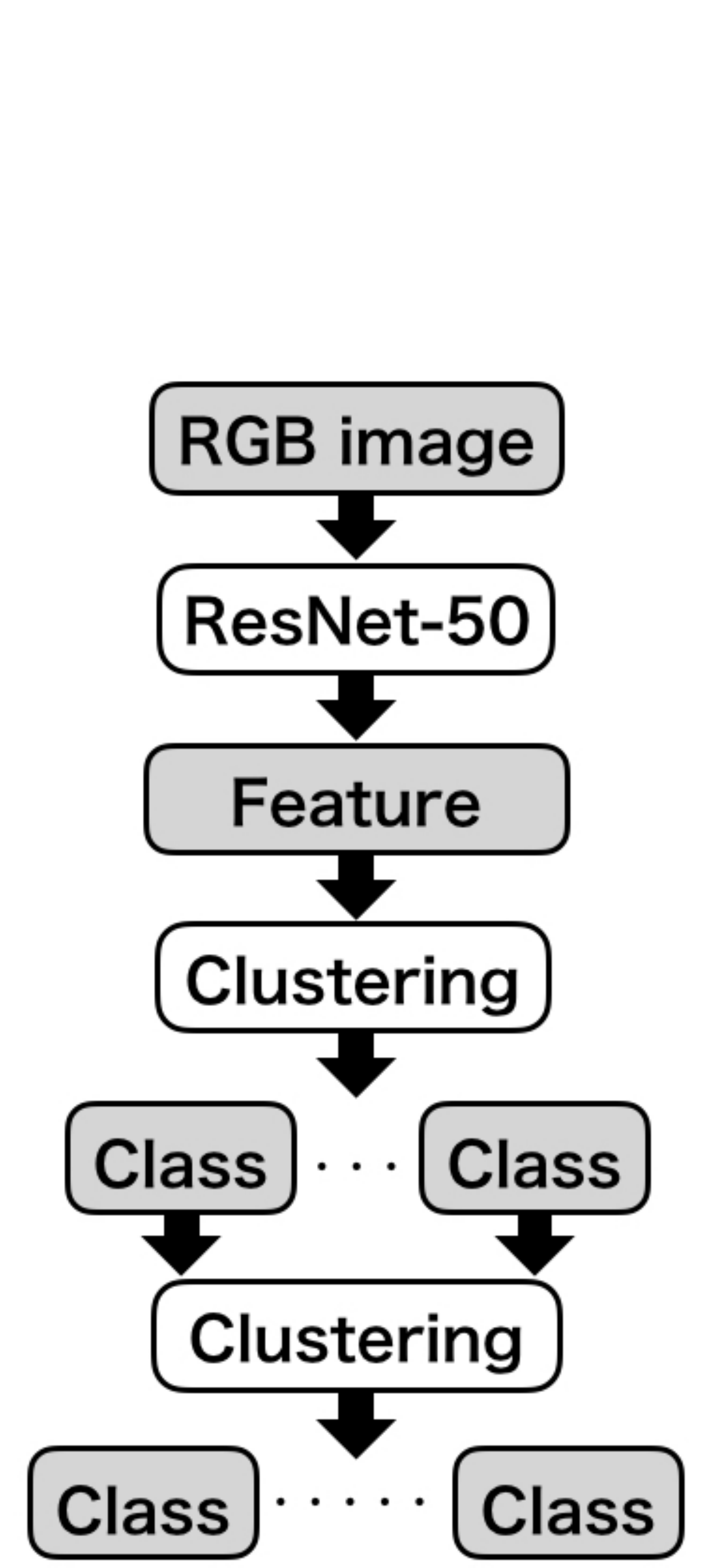}
	\label{fig:rgb_res_cluster}
	}
	\hspace{-7pt}
	\subfigure[Edge\_ResNet\_dataset]{ 
	\includegraphics[width = 2.9cm]{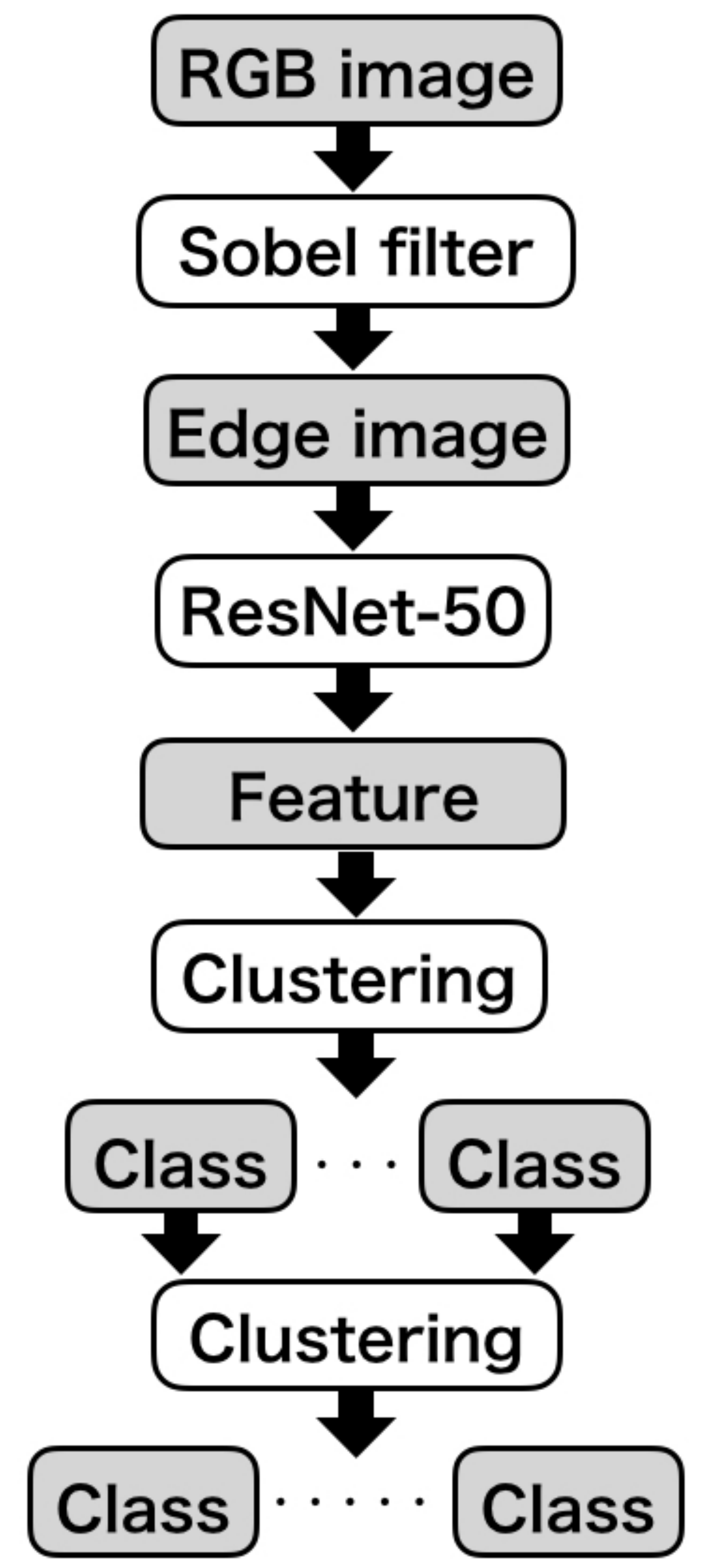}	
	\label{fig:edge_res_cluster}	
	}
	\end{tabular}
\caption[ResNet clustering]{Labeling by clustering}
\vspace{0.3cm}
\label{fig:res_cluster}
\end{figure}

We propose to use a clustering method based on feature extraction using deep learning as a relatively robust method for such problems. Specifically, we used ResNet-50\cite{he2016deep} trained on ImageNet. Using the output of the average pooling layer of the ResNet-50 as a feature from the sample of the basic set, the number of clusters was estimated from this feature, as shown in Section \ref{sssec:no}. Then, using K-means++, the \textit{yuru-chara} image dataset was divided by the number of classes. This process is called ResNet clustering. In the current paper, we investigated whether each class obtained by ResNet clustering can be further divided by applying ResNet clustering again. The flow of this clustering method is shown in \wfig{rgb_res_cluster}. The label obtained by this method was added to all the samples in the \textit{augmented set}, and the resulting dataset is named the \texttt{RGB\_ResNet\_dataset}. The underlying idea behind using ResNet is that the K-means++ has a problem that the clustering result varies greatly depending on the slide or rotation of the image. When clustering by the K-means++ using the samples as they are, the characters affected by this problem may be classified into another class, even if they have similar features. 
 
 \subsubsection{Edge ResNet Feature}
 \label{sec:edge_resnet}
 In addition, we used the edge image extracted by a Sobel filter\cite{kanopoulos1988design} for ResNet clustering and converted it to a labeled dataset in the same flow. This is named the \texttt{Edge\_ResNet\_dataset}. The reason for using edge images is that deep cluster\cite{Caron_2018_ECCV} has high clustering performance when using edge images. For characters, clustering can be performed based on information such as silhouettes and poses.

\subsection{Image Generation Method}
\subsubsection{Image Generative Model}
\begin{figure}[t]
	\centering
	\begin{tabular}{c}
	\subfigure[ResBlock for the generator]{
	\includegraphics[width = .8\columnwidth]{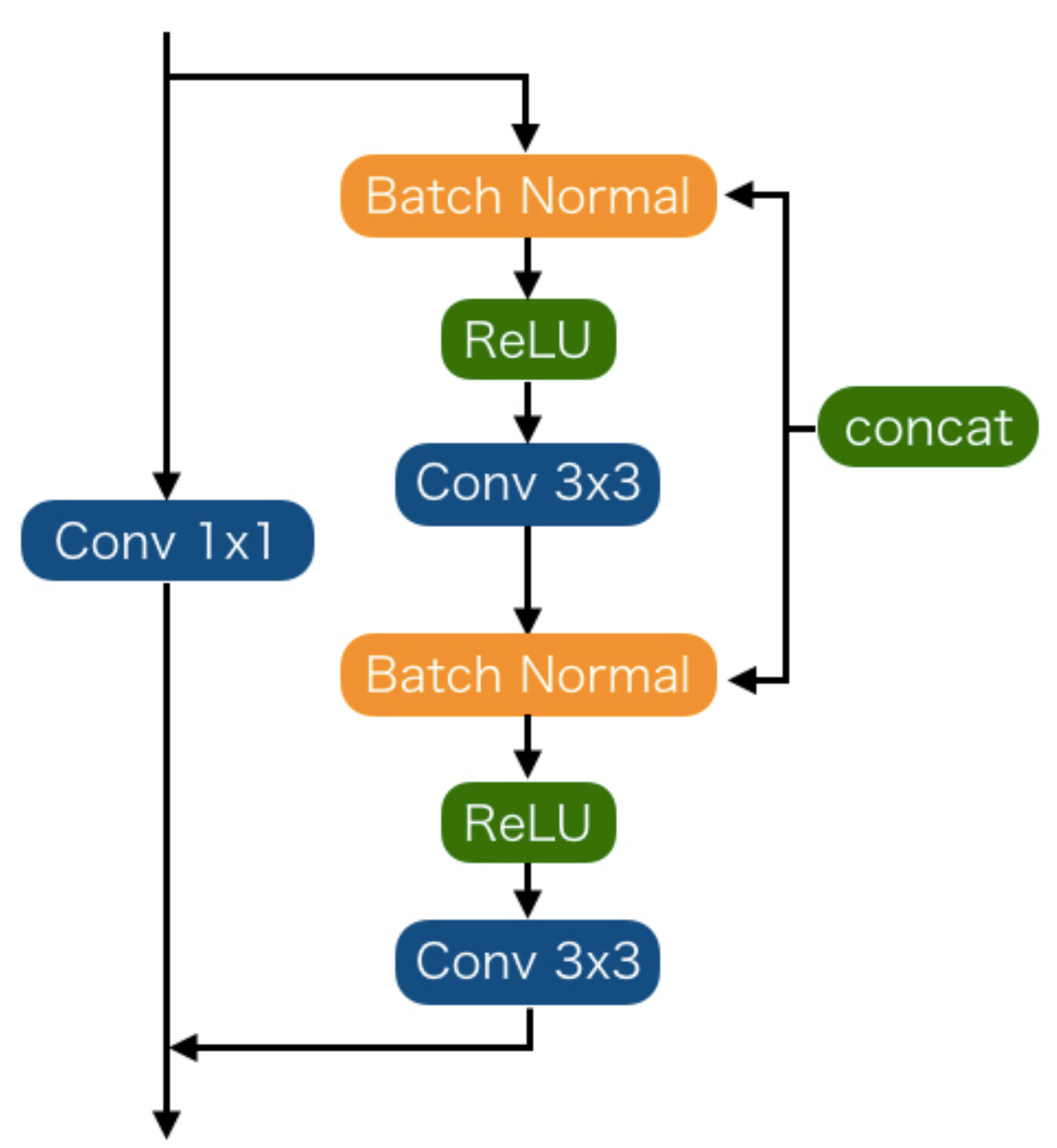}
	\label{fig:res_g}
	}
	\\
	\subfigure[ResBlock for the discriminator]{ 
	\includegraphics[width = .8\columnwidth]{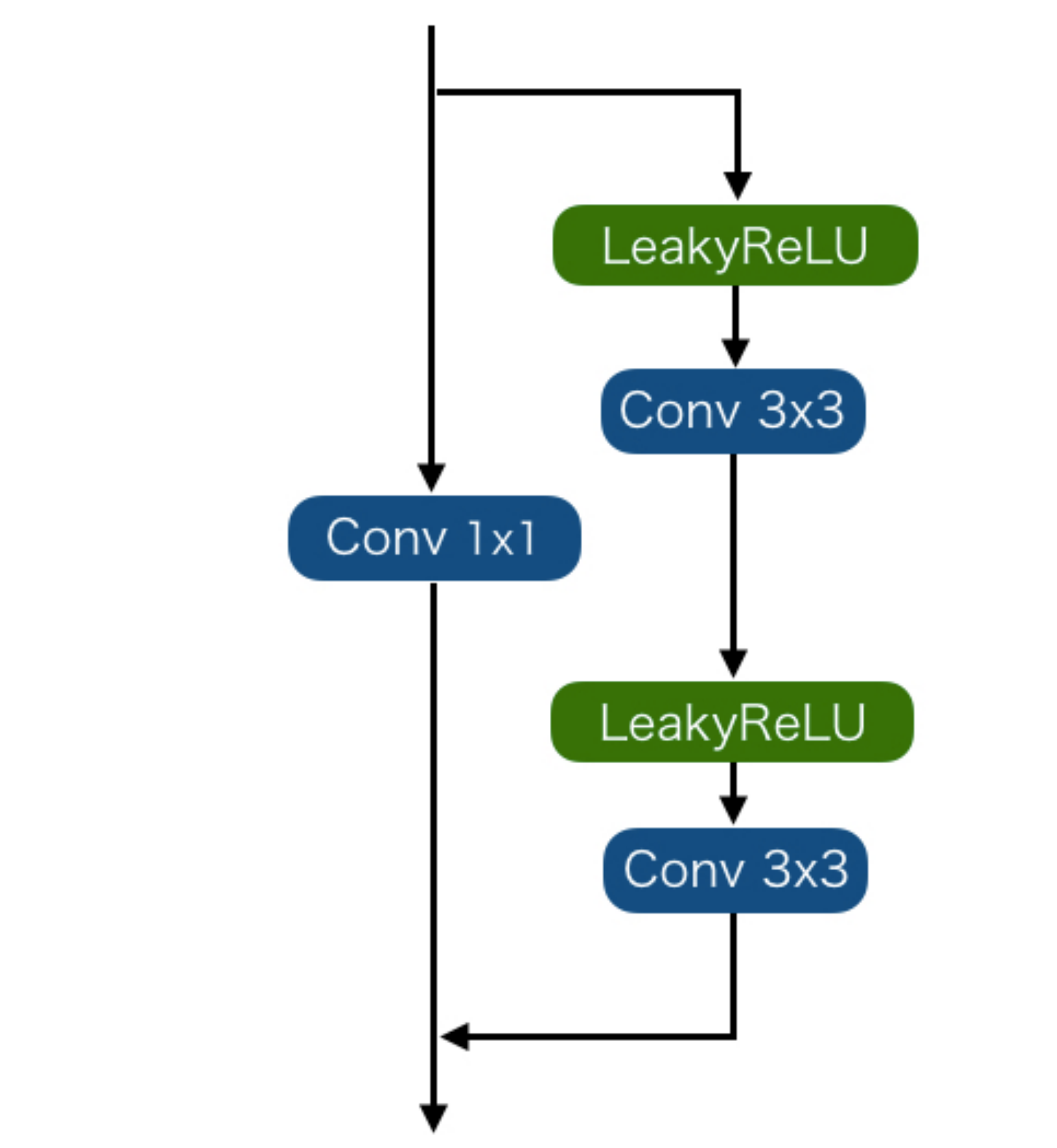}	
	\label{fig:res_d}	
	}
	\end{tabular}
	\caption[ResBlock ]{Architecture of ResBlocks}
\label{fig:fig4}
\end{figure}

The proposed image generative model is based on an auxiliary classifier GAN (AC-GAN)\cite{odena2017conditional}, and spectral normalization\cite{miyato2018spectral} and self-attention layer\cite{zhang2018self} are applied. Moreover, ResBlock\cite{he2016deep, miyato2018spectral} is used for convolution, as illustrated in (\wfig{fig4}).
    
\begin{figure}[t]
　\centering
　\includegraphics[width=1.3\columnwidth, angle=-90]{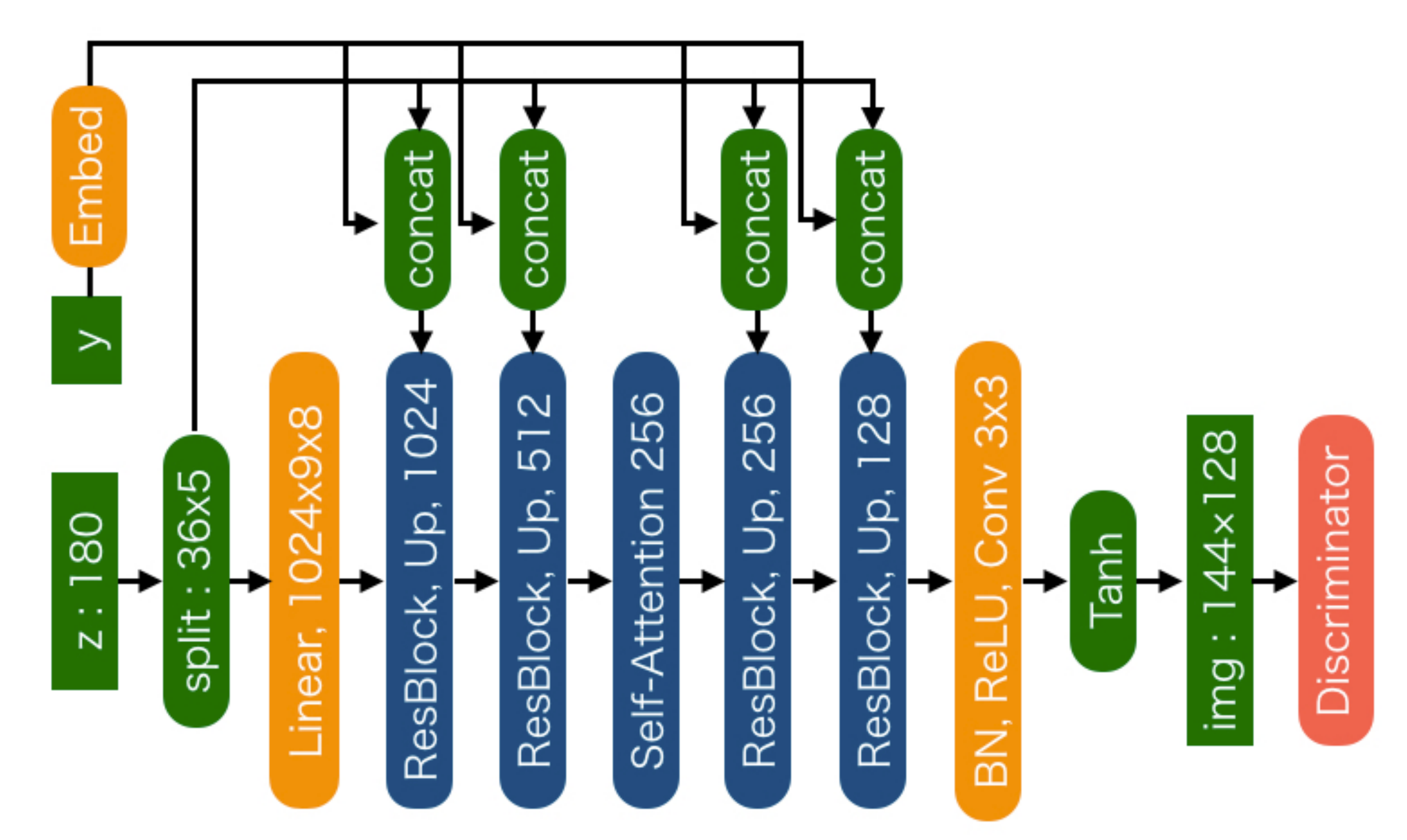}
　\caption{Architecture of the generator}
　\label{fig:fig5}
\end{figure}

The architecture of the generator is shown in \wfig{fig5}. In the generator, a class label was assigned to each ResBlock by conditional batch normalization\cite{miyato2018spectral}. Latent noise $z$ was divided and given to ResBlock step by step by hierarchical latent space\cite{brock2018large}. The structure of ResBlock for the generator is shown in \wfig{res_g}.

\begin{figure}[t]
　\centering
　\includegraphics[width=1.3\columnwidth, angle=-90]{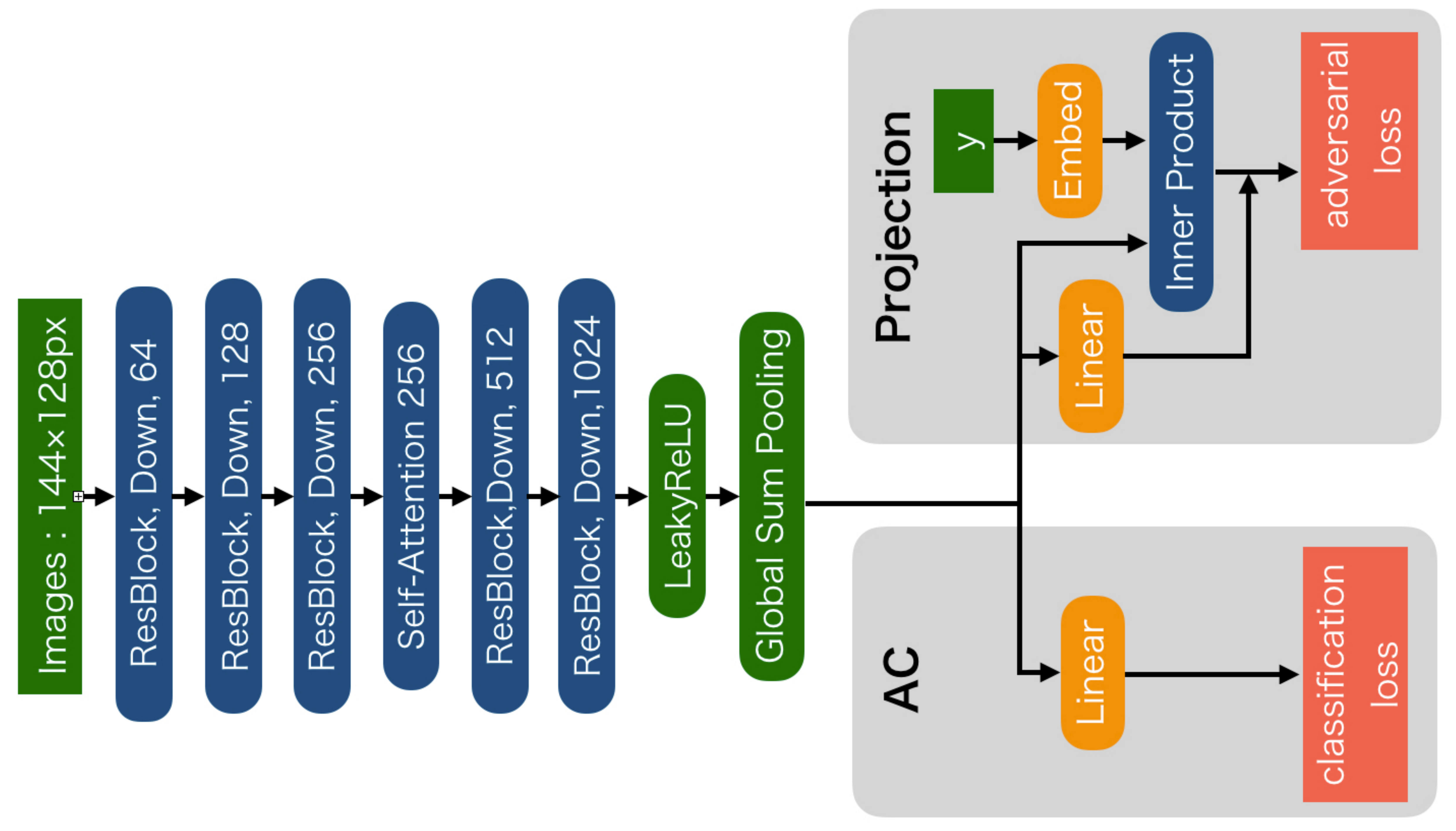}
　\caption{Architecture of the discriminator}
　\label{fig:fig6}
\end{figure}

The architecture of the discriminator is shown in \wfig{fig6}. Because the discriminator is based on an AC-GAN, an auxiliary classifier (AC)\cite{odena2017conditional} shown in \wfig{fig6} is included. The discriminator is given a class label by projection layer\cite{miyato2018cgans}. The structure of ResBlock for the discriminator is shown in \wfig{res_d}.

\subsubsection{Parameters}
Adam\cite{adam} was used as the optimizer for each generator and discriminator. It has been reported that if the learning rate of the discriminator is larger than the learning rate of the generator, then the convergence to the Nash equilibrium and the model performance will be improved. Therefore, based on this argument, the learning rate of the discriminator was set to $5\cdot10^{-4}$, and the learning rate of the generator was set to $2\cdot10^{-5}$. The model was trained for 500 epochs with a batch size of 32. At this time, the number of updates of the discriminator was set to twice that of the generator, and the discriminator was trained for 627K iterations and the generator for 313.5K iterations.

\subsection{Evaluation Index}
In the present paper, we use the geometry score (GS) \cite{khrulkov2018geometry} proposed by Khrulkov and Oseledets as an evaluation index of model performance and the generated images. The GS is based on the manifold hypothesis that data in a high-dimensional space can be inclined to manifolds in the lower-dimensional space, here considering GAN training data and generated images as manifolds. By using the GS, it is possible to check how close the properties of the generated image are to the training data and to evaluate how much the model can learn from the training data. However, because the number of training data and generated images is very large, it is difficult to calculate them as manifolds. Therefore, in the GS, the training data and the generated image are considered as a witness complex\cite{witness2004} by randomly sampling the data points. By using the witness complex, the mean survival time (mean relative living time; MRLT) of one-dimensional holes on the manifold can be calculated. The GS can be obtained by calculating the MRLT from the training data and the generated image and then by taking the square error. The smaller the GS, the closer the properties of the generated image are to the properties of the training data, indicating that the performance of the model is high.

\section{Results}
\label{ch:result}
\subsection{Clustering Result}
We performed three types of clustering. Details of the dataset obtained by clustering are shown in \wtab{clustering}. In the clustering method shown in \wfig{rgb_clustering}, the \textit{yuru-chara} image dataset was divided into 10 classes. In the clustering method shown in \wfig{rgb_res_cluster}, the character image dataset was divided into 11 classes in the first ResNet clustering and 12 classes in the second ResNet clustering. In the clustering method shown in \wfig{edge_res_cluster}, the character image dataset was divided into 12 classes in the first ResNet clustering and 16 classes in the second ResNet clustering.

 \begin{table}[t]
\caption{Details of the dataset obtained by clustering}
\centering
\scalebox{0.94}{
\begin{tabular}{|c|c|c|} \hline
{dataset} & {feature} & {number of class} \\ \hline
{\texttt{RGB\_dataset}}&{RGB image}&{10}\\ \hline
{\texttt{RGB\_ResNet\_dataset}}&{RGB image + ResNet-50}&{12}\\ \hline
{\texttt{Edge\_ResNet\_dataset}}&{edge image + ResNet-50}&{16}\\ \hline
\end{tabular}
}
\vspace{0.3cm}
\label{tab:clustering}
\end{table}

\subsection{Learning Result}
\subsubsection{Generated Image}
\label{sec:gen_result}
\begin{figure*}[h]
	\centering
	\begin{tabular}{cc}
	\subfigure[\texttt{RGB\_dataset}]{
	\includegraphics[width = .3\textwidth]{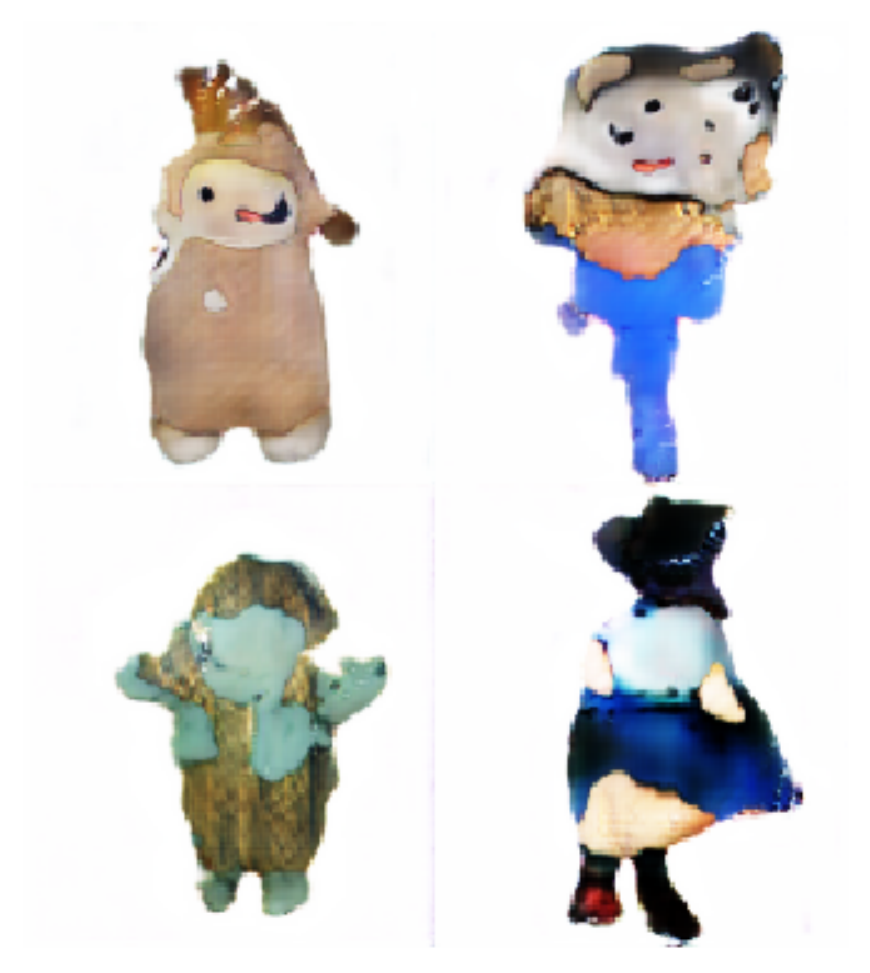}
	\label{fig:conp150}
	}
	\subfigure[\texttt{RGB\_ResNet\_dataset}]{
	\includegraphics[width = .3\textwidth]{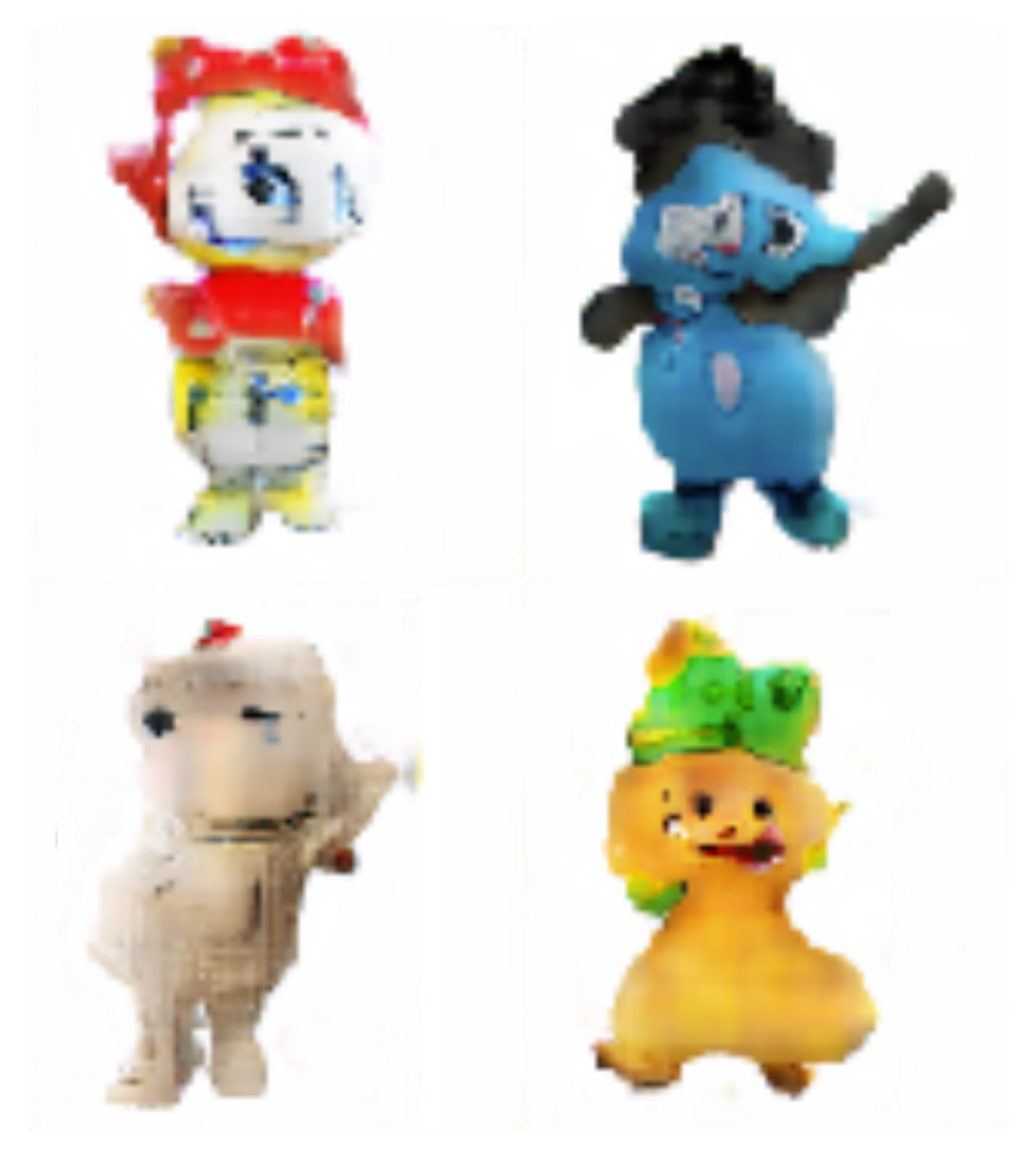}
	\label{fig:rr150}
	}
	\subfigure[\texttt{Edge\_ResNet\_dataset}]{ 
	\includegraphics[width = .3\textwidth]{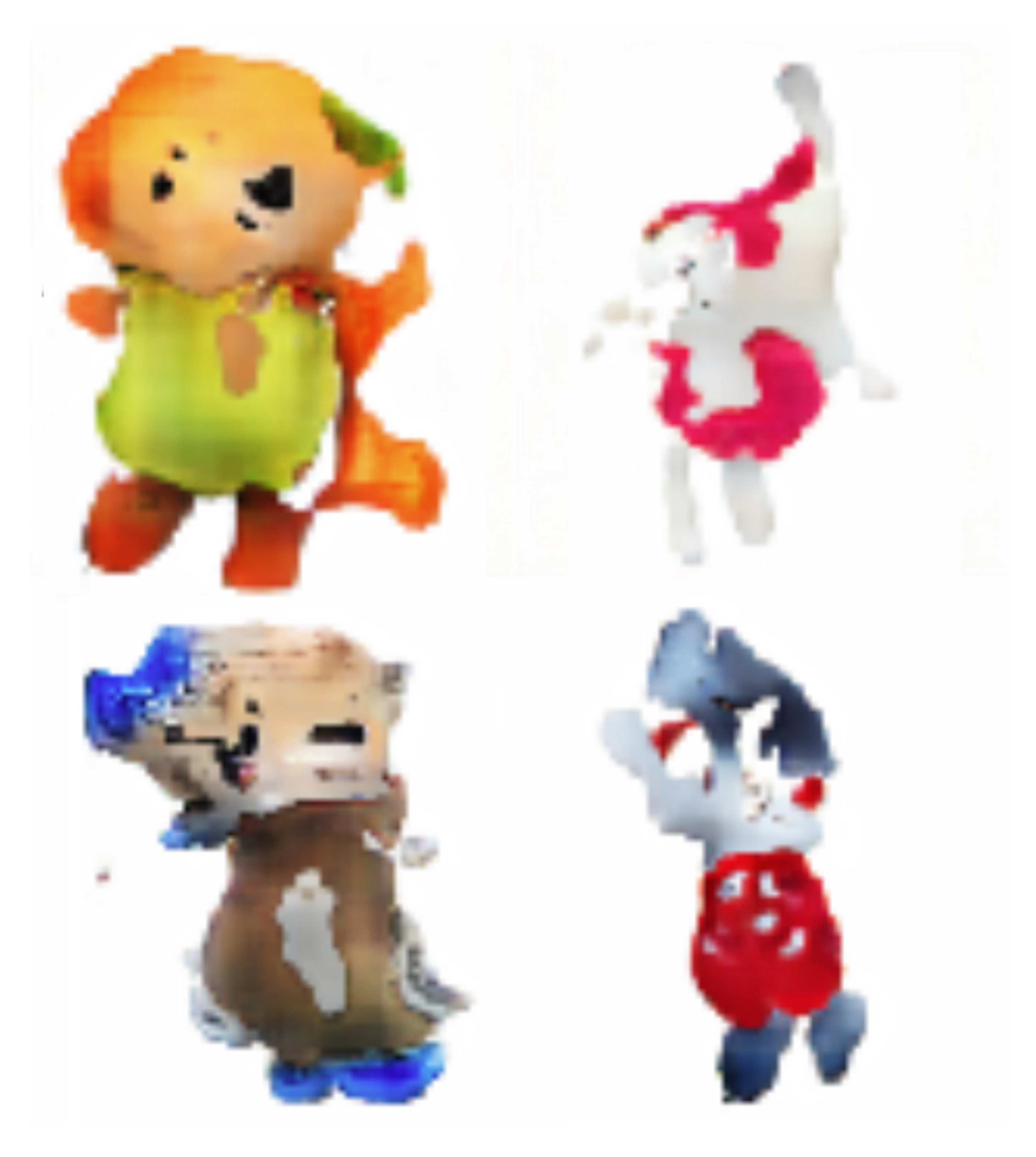}	
	\label{fig:er150}	
	}
	\end{tabular}
\caption[ResNet clustering]{Generated image (epoch：150)}
\label{fig:syn150}
\end{figure*}

\begin{figure*}[h]
\centering
	\begin{tabular}{cc}
	\subfigure[\texttt{RGB\_dataset}]{
	\includegraphics[width = .3\textwidth]{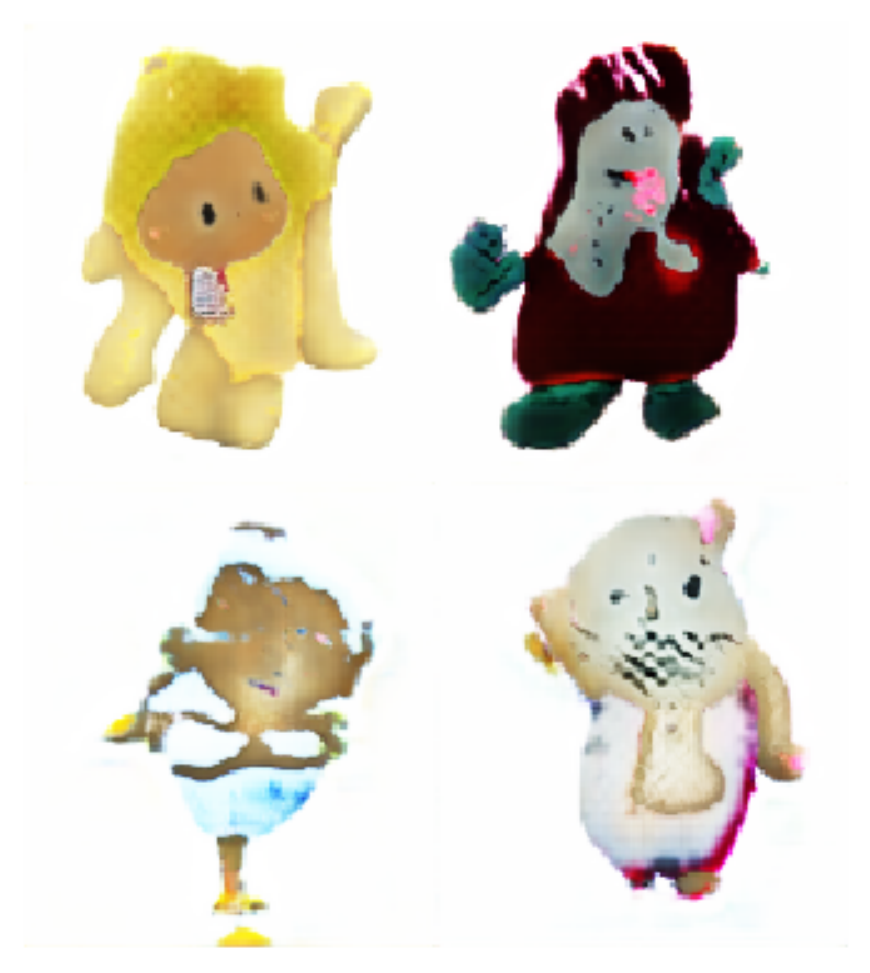}
	\label{fig:conp500}
	}
	\subfigure[\texttt{RGB\_ResNet\_dataset}]{
	\includegraphics[width = .3\textwidth]{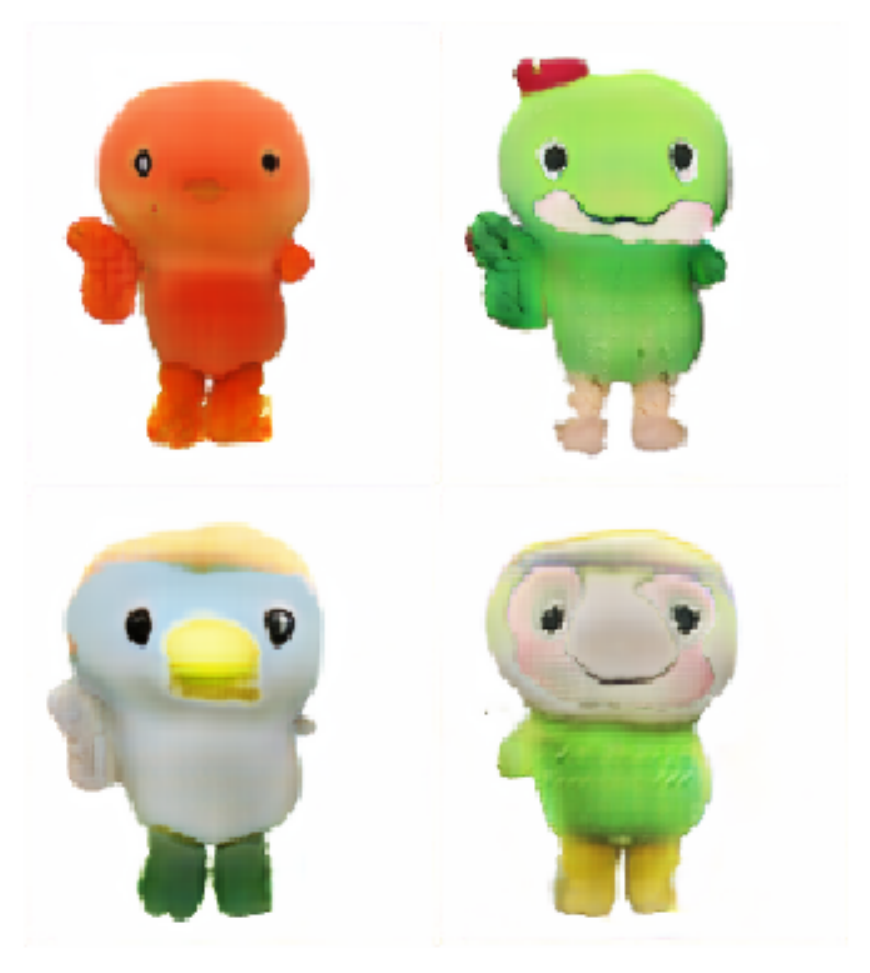}
	\label{fig:rr500}
	}
	\subfigure[\texttt{Edge\_ResNet\_dataset}]{ 
	\includegraphics[width = .3\textwidth]{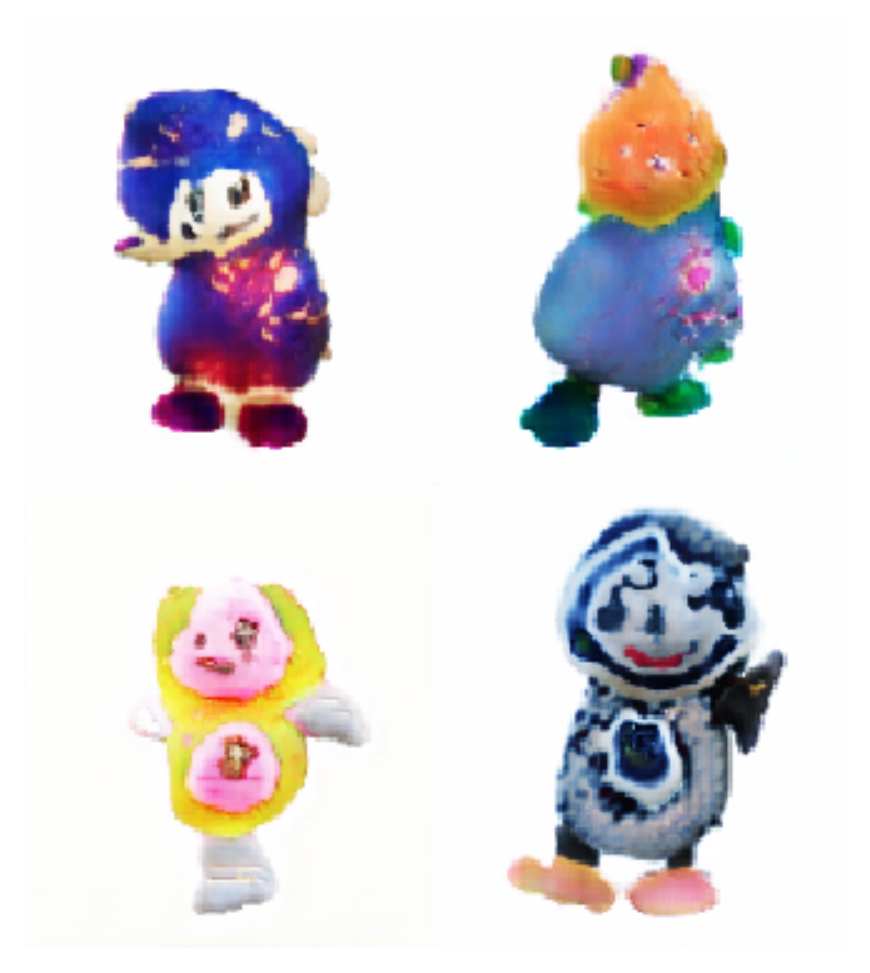}	
	\label{fig:er500}	
	}
	\end{tabular}
\caption[ResNet clustering]{Generated image (epoch：500)}
\label{fig:syn500}
\end{figure*}

\begin{figure*}[t]
\centering
	\begin{tabular}{cc}
	\subfigure[\texttt{RGB\_dataset}]{
	\includegraphics[width = \textwidth]{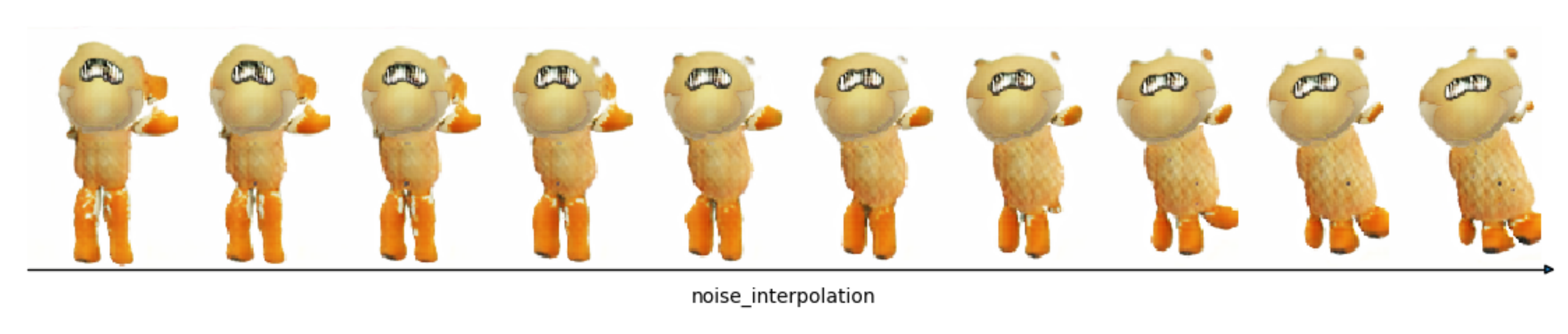}
	\label{fig:rgb_nter}
	} \\
	\subfigure[\texttt{RGB\_ResNet\_dataset}]{
	\includegraphics[width = \textwidth]{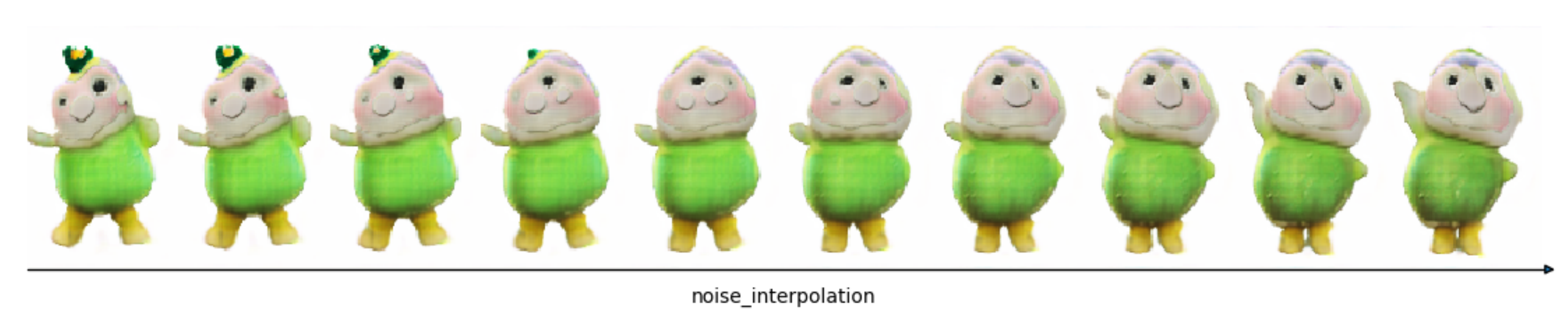}
	\label{fig:rr_inter}
	} \\
	\subfigure[\texttt{Edge\_ResNet\_dataset}]{ 
	\includegraphics[width = \textwidth]{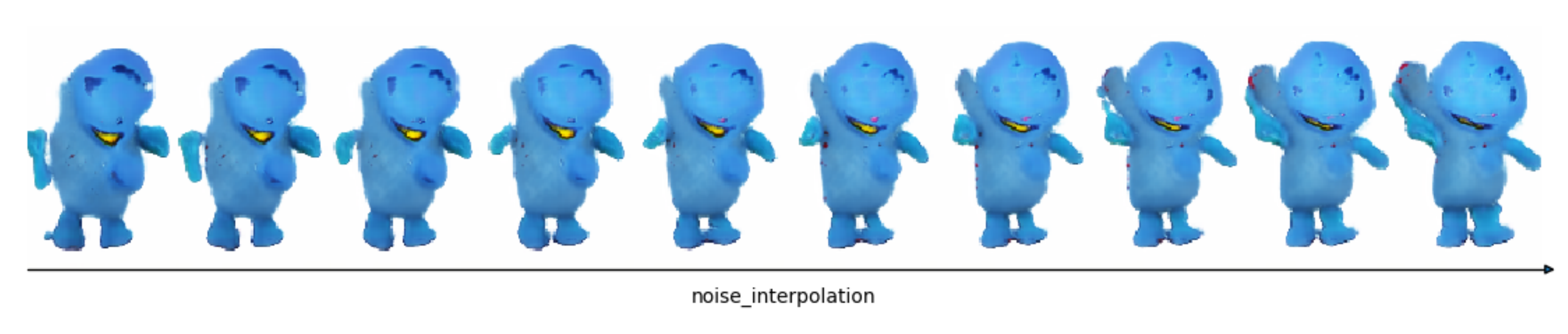}	
	\label{fig:er_inter}	
	}
	\end{tabular}
\caption[latent noise interpolation]{Generated image when latent noise is gradually interpolated with the input class fixed.}
\label{fig:interpolation}
\end{figure*}

\begin{table}[t]
 \caption{GS $\times 10^3$ of the generated image obtained from the model trained on each dataset}
  \centering
  \begin{tabular}{|c|c|c|} \hline
   {\texttt{RGB\_dataset}}& {\texttt{\texttt{RGB\_ResNet\_dataset}}} & {\texttt{Edge\_ResNet\_dataset}} \\ \hline
  {12.8}&{\textbf{5.5}}&{15.5}\\ \hline
  \end{tabular}
 \label{tab:gs}
\end{table}

 \begin{figure}[t]
　\centering
　\includegraphics[width=\columnwidth]{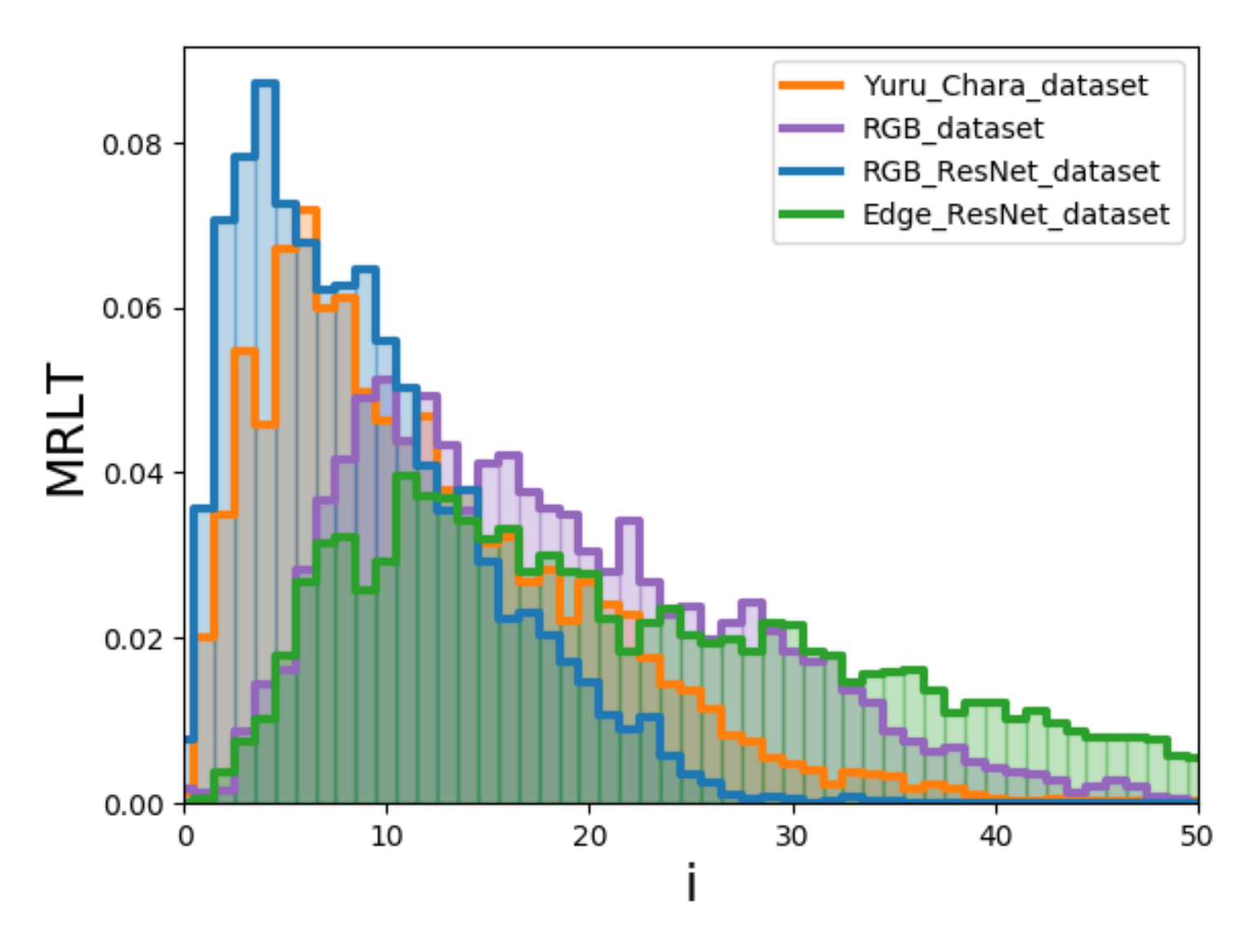}
　\caption{Distribution of mean relative living time (MRLT)}
　\label{fig:gsbar}
\end{figure}

We trained models using the datasets obtained by three types of clustering methods: \texttt{RGB\_dataset}, \texttt{RGB\_ResNet\_dataset}, and \texttt{Edge\_ResNet\_dataset}. In Figures 7(a) to (c), the generated images of the model trained 150 epochs for each dataset are shown. As long as these figures are compared visually, there is no significant difference in the quality of the generated images from differences in the datasets. From Figures 8(a) to (c), the generated images of the model that trained 500 epochs for each dataset are shown. From this generation result, it was found that the model trained with \texttt{RGB\_ResNet\_dataset} could generate the character by capturing characteristics such as the silhouette, eyes, and mouth of the character. Also, comparing Figures 7(b) and 8(b), when the epoch increases from 150 to 500, the generated characters show the features of the eyes and mouth, and as the training progresses, training using \texttt{RGB\_ResNet\_dataset} is performed. It was found that the model learned the characteristics of the character well.

\wfig{interpolation} shows the result of interpolating the latent noise by 10 steps, using the latent noise of each step as the input, and generating an image using a model trained on each dataset. As seen in \wfig{interpolation}, in any dataset, the model, such as the style and pose of the generated character, changes smoothly with the change of the input noise.

\subsubsection{Evaluation by the Geometry Score}
\label{sec:gs_result}
\wtab{gs} shows the results of evaluating the images generated using the three types of datasets by the GS. As a result of evaluating the generated image using the GS, it was found that the model trained with \texttt{RGB\_ResNet\_dataset} could generate an image with properties closest to the training dataset. With the horizontal axis representing the number of one-dimensional holes $i$ and the vertical axis representing MRLT, the distribution of the MRLT of the generated image of the model trained in each dataset and the \textit{yuru-chara} image dataset is shown in \wfig{gsbar}. As seen in \wfig{gsbar}, the MRLT distribution of the generated image using \texttt{RGB\_ResNet\_dataset} is close to the MRLT distribution of the training data, so it can be said that images with properties similar to the training data can be generated without mode collapse. In the model trained with \texttt{RGB\_dataset} or \texttt{Edge\_ResNet\_dataset}, the mountain of the MRLT distribution of the generated image is greatly spread to the left and right, indicating that the properties of the training data were not well learned.

\section{Discussion}
\label{ch:discussion}
\subsection{Quality of Generated Image}
From the results of the \ref{sec:gen_result}, clustering using ResNet-50\cite{he2016deep} is effective for improving the quality of the generated image. This is because clustering using ResNet-50 can provide effective class information from the \textit{yuru-chara} image dataset, and the model is trained using the class information, which has led to efficient learning.  As shown in Figures 8(b) and 8(c), the quality of the generated image in the model trained with \texttt{Edge\_ResNet\_dataset} is lower than that trained with \texttt{RGB\_ResNet\_dataset}. The reason for this is that the ResNet-50 used in the current paper was trained with ImageNet\cite{deng2009imagenet}, which is not a collection of edge-extracted images. Therefore, this suggests that the clustering method using edge images did not yield a dataset that allowed the proposed model to be successfully trained.

\subsection{Distribution of Generated Image}
As described in Section \ref{sec:gs_result}, the generated image was evaluated by GS, and it was found that the model using \texttt{RGB\_ResNet\_dataset} was able to generate an image that most closely resembled the character image dataset. The reason for this is because ResNet can be used for clustering to divide the loose character image dataset into an appropriate number of classes, so the model can efficiently learn the properties of the dataset from the class information obtained by the clustering. Furthermore, the model trained with \texttt{Edge\_ResNet\_dataset} has a higher GS than that with \texttt{RGB\_ResNet\_dataset}. The highest GS for \texttt{Edge\_ResNet\_dataset} can be explained by the number of classes determined by the clustering. The number of classes with the Edge ResNet dataset is the largest among the three datasets that were simulated in this paper. It has been reported that when it comes to the AC-GAN\cite{odena2017conditional}, too large number of classes can lead to unstable learning. The model proposed in the current paper is also based on AC-GAN. In the original paper on AC-GAN, the upper limit of the number of classes was set to 10. In our experiment, the number of classes in the \texttt{Edge\_ResNet\_dataset} was 16, which far outweighs this number. This may suggest that the learning was not stable and that the proposed model did not sufficiently learn the properties of the dataset. 

\section{Conclusion}
In the current paper, we applied the X-means and the K-means++  to the \textit{yuru-chara} image dataset and created a dataset with classes for comparison. In addition, clustering based on ResNet and edge extraction  and features extracted by ResNet from \textit{yuru-chara} images was performed to create a dataset with classes. By giving these datasets to a class conditional GAN model, a \textit{yuru-chara} image was generated. As a result, when ResNet was used for clustering, it was possible to generate high-quality \textit{yuru-chara} images that were close to the characteristics of the \textit{yuru-chara} image dataset. The results suggest that clustering of datasets based on features obtained by ResNet may be effective for improving the quality of images generated by GANs when the dataset is small. In the future, it is necessary to investigate whether clustering by ResNet is effective for other small datasets.

\section*{Acknowledgment}
This work was supported in part by JSPS KAKENHI, 17H01760.

\bibliographystyle{IEEEbib}
\bibliography{ref}

\end{document}